\documentclass[sigconf]{aamas} 

\renewcommand\footnotetextcopyrightpermission[1]{} 
\usepackage{balance} 

\usepackage{times}
\usepackage[ruled, vlined, linesnumbered]{algorithm2e}

\usepackage{hyperref}
\usepackage{url}
\usepackage{graphicx}
\usepackage{subcaption}
\graphicspath{ {./img/} }

\usepackage{mathtools}
\usepackage{afterpage}

\DeclarePairedDelimiterX{\infdivx}[2]{(}{)}{%
  #1\;\delimsize\|\;#2%
}
\newcommand{\kl}{KL\infdivx}

\newcommand{\R}{\mathbb{R}}

\newcommand{\seq}[1]{\left\langle #1 \right\rangle}

\newcommand{\set}[1]{\left\{ #1 \right\}}

\setcopyright{ifaamas}
\acmConference[AAMAS '21]{Proc.\@ of the 20th International Conference on Autonomous Agents and Multiagent Systems (AAMAS 2021)}{May 3--7, 2021}{London, UK}{U.~Endriss, A.~Now\'{e}, F.~Dignum, A.~Lomuscio (eds.)}
\copyrightyear{2021}
\acmYear{2021}
\acmDOI{}
\acmPrice{}
\acmISBN{}

\acmSubmissionID{???}

\title{Semi-On-Policy Training for Sample Efficient Multi-Agent Policy Gradients}

\author{Bozhidar Vasilev}
\affiliation{
  \department{Department of Computer Science}
  \institution{University of Oxford}}
\email{bozhidar.vasilev1@gmail.com}

\author{Tarun Gupta}
\affiliation{
  \department{Department of Computer Science}
  \institution{University of Oxford}}
\email{tarun.gupta@cs.ox.ac.uk}

\author{Bei Peng}
\affiliation{
  \department{Department of Computer Science}
  \institution{University of Oxford}}
\email{bei.peng@cs.ox.ac.uk}

\author{Shimon Whiteson}
\affiliation{
  \department{Department of Computer Science}
  \institution{University of Oxford}}
\email{shimon.whiteson@cs.ox.ac.uk}

\begin{abstract}
Policy gradient methods are an attractive approach to multi-agent reinforcement learning problems due to their convergence properties and robustness in partially observable scenarios. However, there is a significant performance gap between state-of-the-art policy gradient and value-based methods on the popular StarCraft Multi-Agent Challenge (SMAC) benchmark. In this paper, we introduce semi-on-policy (SOP) training as an effective and computationally efficient way to address the sample inefficiency of on-policy policy gradient methods. We enhance two state-of-the-art policy gradient algorithms with SOP training, demonstrating significant performance improvements. Furthermore, we show that our methods perform as well or better than state-of-the-art value-based methods on a variety of SMAC tasks.

\end{abstract}


\newcommand{\BibTeX}{\rm B\kern-.05em{\sc i\kern-.025em b}\kern-.08em\TeX}

\begin{document}


\pagestyle{fancy}
\fancyhead{}


\maketitle 


\section{Introduction}

Many real-world learning problems, such as autonomous vehicle coordination \citep{autonomous_vehicles} and network packet delivery \citep{wireless_routing}, can be naturally modelled as cooperative multi-agent reinforcement learning (MARL) problems. Reinforcement learning (RL) algorithms designed for single-agent settings are typically unsuitable for multi-agent scenarios, because the joint action space grows exponentially with the number of agents. Furthermore, even when the joint action space is not prohibitively large, communication constraints may make it impossible to apply single-agent algorithms directly. 

This often makes it necessary to learn \textit{decentralised} policies, where the agents make decisions independently, based only on their own local observations. However, if the agents learn independently, this makes the environment non-stationary from the viewpoint of any agent, so learning might not converge.
To address this issue, one often adopts the framework of \textit{centralised training with decentralised execution} (CTDE) \citep{oliehoek2008optimal,kraemer2016}, assuming a simulated or controlled setting. The CTDE framework allows the algorithm to use any extra information available during training. However, at test time each agent must make decisions based only on its private observations.

The StarCraft Multi-Agent Challenge (SMAC) \cite{smac} is a challenging MARL benchmark in which agents must learn complex coordination behaviours in an environment with significant partial observability. In recent years SMAC has become a popular environment for evaluating cooperative MARL algorithms \cite{wqmix,mahajan2019maven,wang2020dop,zhou2020learning}.
Despite the progress in multi-agent policy gradient (MAPG) methods, most of them still struggle to compete with state-of-the-art value-based methods on SMAC.
\citet{qmix_journal} show that COMA \citep{coma}, a state-of-the-art on-policy MAPG algorithm, severely underperforms even simple value-based methods. 
\citet{papoudakis2020comparative} go as far as using $10$ times more data for MAPG methods when comparing against value-based ones. They show that state-of-the-art on-policy MAPG methods, such as CentralV \cite{coma} and COMA, suffer from poor performance even on simple SMAC scenarios. 

While off-policy algorithms are generally more sample efficient than on-policy ones, state-of-the-art off-policy MAPG methods, such as MADDPG \citep{lowe2017maddpg} and Multi-Agent Soft Actor-Critic (MASAC)
\citep{haarnoja2018soft,masac}, also perform poorly on SMAC \citep{wqmix}. The recently proposed stochastic decomposed policy gradient (DOP) \cite{wang2020dop} algorithm mixes off-policy and on-policy training, by adapting the tree backup technique \citep{eligibility_traces, NIPS2016_c3992e9a} to the multi-agent setting and taking advantage of a factored critic representation. The authors claim their method is the first MAPG algorithm that outperforms state-of-the-art value-based methods on SMAC. However, they only present results for a small number of SMAC maps.

In this paper, we improve COMA by fixing a representation issue with the centralised critic, which produces inconsistent value estimates for some joint actions due to its network architecture. We further improve COMA by changing its training scheme to use the entire batch of data, rather than minibatches.
We find that our version, COMA with a consistent critic (COMA-CC), consistently outperforms the original COMA on SMAC, and provide ablation studies to evaluate the effect of each change.

Furthermore, we introduce semi-on-policy (SOP) training as an effective and computationally efficient way to address the sample inefficiency of on-policy policy gradient algorithms. Key to our method is using recent off-policy data for training, based on an eligibility criterion. We apply SOP training to CentralV and COMA-CC, and demonstrate significant performance improvements. Our methods achieve similar or better performance and sample efficiency, compared to state-of-the-art value-based methods, on a variety of challenging SMAC tasks. Additionally, our best performing method often outperforms other state-of-the-art MAPG methods, including DOP.

\section{Related Work}
\label{sec:related_work}
\paragraph{Value-based methods} A natural value-based MARL method for learning decentralised policies is independent $Q$-learning \cite{tan1993multi}, where each agent learns an individual action-value function and acts based on it; this is typically done using DQN \cite{Mnih2015,maiql}. To exploit CTDE, \citet{vdn} propose value decomposition networks (VDN), where a centralised but factored action-value function is learned, which is the sum of the individual agent utilities. The QMIX algorithm \cite{qmix_journal} expands this idea by learning a factored joint action-value function that is only constrained to be monotonic with respect to each agent's utility. Recent works, like Weighted-QMIX \cite{wqmix} and Qatten \cite{qatten}, aim to address the limitations imposed by the monotonicity constraint of QMIX, while still using a monotonic factorisation. On the other hand, works like QTRAN \cite{qtran} address the monotonic restriction of QMIX by learning a VDN-factored joint action-value function along with an unrestricted centralised state-value function,
and QTRAN++ \cite{qtran++} address the gap between the theoretical and empirical results of QTRAN. 

\paragraph{Policy gradient methods} The most successful policy gradient methods have been actor-critic methods, where both a policy and a value function are learned during training, and the estimates of the value function are used to train the policy \cite{sutton}. If each agent learns independently, we get the independent actor-critic (IAC) algorithm, which is an adaptation of the advantage actor-critic (A2C) \cite{a3c} algorithm  to the multi-agent setting. The CentralV algorithm \cite{coma} is another adaptation of A2C, but using a centralised critic that makes use of global state information for learning. To address the issue of multi-agent credit assignment, the COMA algorithm \cite{coma} uses a counterfactual baseline, which marginalises out an agent's action when computing the advantage of their action. \citet{zhou2020learning} address the same problem with their technique of learning implicit credit assignment (LICA) by formulating a centralised critic as a hypernetwork. To improve sample efficiency, the stochastic decomposed policy gradients algorithm (DOP) \cite{wang2020dop} uses a mixture of on-policy and off-policy data by adapting the tree backup technique \citep{eligibility_traces, NIPS2016_c3992e9a} to the multi-agent setting and taking advantage of a factored critic representation. \citet{masac} adapt the soft actor-critic (SAC) \cite{haarnoja2018soft} algorithm to the multi-agent setting and combine it with an attention mechanism and a centralised critic. The MADDPG \cite{lowe2017maddpg} algorithm is an adaptation of the DDPG algorithm \cite{lillicrap2019ddpg} to the multi-agent setting; while originally the algorithm was used to learn deterministic policies in continuous domains using off-policy data, it has since been adapted to discrete domains as well \cite{wqmix, masac}.  

\paragraph{SMAC} The StarCraft Multi-Agent Challenge (SMAC) \citep{smac} is a popular cooperative multi-agent reinforcement learning benchmark. The challenges require agents to learn complex cooperative behaviours, such as \textit{kiting} and \textit{focus fire}. For a long time, value-based methods have significantly outperformed policy gradient methods on SMAC. Some recent works, such as LICA \cite{zhou2020learning} and DOP \cite{wang2020dop} have shown that actor-critic methods can sometimes outperform value-based methods on SMAC. However, they only focus on a limited set of maps, and their algorithms are much more complex compared to previous work.

Our work focuses on improving the sample efficiency of on-policy policy gradient algorithms by directly using off-policy data during training, based on an eligibility criterion. In principle, this makes it applicable to any such method. In contrast to methods that correct for off-policy data using importance sampling or returns corrections \cite{acer, retrace, wang2020dop,eligibility_traces,qlambda_offpolicy, pmlr-v70-foerster17b}, we simply require the generating policy for all training data to be closely related to the current behaviour policy. This removes the associated computational cost of applying corrections to the data. Furthermore, in contrast to proximal policy optimisation (PPO) \cite{ppo}, which uses a "surrogate" objective to constrain policy updates over multiple training iterations on a fixed set of on-policy data, our method optimises the true objective of the base algorithm and uses a rolling-window replay buffer that contains recent off-policy data, along with on-policy data.
In contrast to \cite{zhou2020learning,wang2020dop}, we show that a simple base algorithm like CentralV suffices for many challenging SMAC tasks, with our method of training. 

\section{Background}
\label{sec:background}
\paragraph{Dec-POMDPs} We consider a \textit{fully cooperative multi-agent task}, which can be modelled as a \textit{Decentralised Partially Observable Markov Decision Process} (Dec-POMDP) \citep{pomdp} consisting of a tuple $G = \seq{S, U, P, r, Z, O, A, \gamma}$. 
The state of the environment at each time step is $s \in S$. Each of $n$ agents $a \in A \equiv \set{1, \dots, n}$ chooses an action $u^a \in U \equiv \set{1, \dots, m}$, forming the joint action $\mathbf{u} \in \mathbf{U} := U^n$. 
The environment dynamics are modelled by a state  transition function $P: S \times \mathbf{U} \times S \rightarrow [0,1]$, where $P(s'|\mathbf{u}, s)$ is the probability that the environment transitions to state $s'$ given that the current state is $s$ and the joint action is $\mathbf{u}$. 
All agents share a global reward function $r: S \times \mathbf{U} \rightarrow \R$. 
We consider partially observable scenarios, where an observation function $O : S \times A \rightarrow Z$ individually determines each agent's observation $z^a$ at each time step. Each agent has a private action-observation history $\tau^a \in T := (Z \times U)^*$, on which it conditions a stochastic policy $\pi^a: U \times T \rightarrow [0,1]$. 
The joint policy $\boldsymbol{\pi}: \mathbf{U} \times T^n \rightarrow [0,1]$ admits a joint state-value function 
$V^{\boldsymbol{\pi}}(s_t) := \mathbb{E}_{\boldsymbol{\pi}}[R_t|s_t]$ 

and action-value function $Q^{\boldsymbol{\pi}}(s_t, \mathbf{u}_t) = \mathbb{E}_{\boldsymbol{\pi}}[R_t|s_t, \mathbf{u_t}]$, where $R_t = \sum_{i=0}^{\infty} \gamma^i r_{t+i}$ is the \textit{discounted return}.

\paragraph{CentralV and COMA}
Both CentralV and COMA \citep{coma} learn decentralised policies $\pi^a$ for each agent $a$, which induce a joint policy $\boldsymbol{\pi}$. During training they maximise the objective $J = V^{\boldsymbol{\pi}}(S_1) = \mathbb{E}_{\boldsymbol{\pi}}[R_1]$, following the policy gradient:
\begin{equation}
   \nabla J = \mathbb{E}_{\boldsymbol{\pi}}\left[\sum_{a}^{}\nabla \log \pi^a(u^a | \tau^a)A_a^{\boldsymbol{\pi}}(s,\mathbf{u})\right],
\end{equation}
where $A_a^{\boldsymbol{\pi}}(s,\mathbf{u})$ is an \textit{advantage} function, which, intuitively, estimates how much better or worse a joint action $\mathbf{u}$ is compared to the agents' expected performance under joint policy $\boldsymbol{\pi}$. As in most cooperative MARL, the agents share policy parameters \citep{qmix_journal, coma,smac}.

In CentralV, the critic approximates the joint state-value function $V_{\psi}(s_t) \approx V^{\boldsymbol{\pi}}(s_t)$, which is used to estimate the advantage for all agents:
\begin{align}
   A_a^{\boldsymbol{\pi}}(s_t, \mathbf{u}_t) &= r_t + \gamma V^{\boldsymbol{\pi}}(s_{t+1}) - V^{\boldsymbol{\pi}}(s_t) \\ 
   &\approx r_t + \gamma V_\psi(s_{t+1}) - V_\psi(s_t)
\end{align}

In COMA, the critic approximates the joint action-value function $Q_\psi(s_t, \mathbf{u}_t) \approx Q^{\boldsymbol{\pi}}(s_t, \mathbf{u}_t)$, 
which is used to estimate a counterfactual advantage for each agent:
\begin{align}
&A^{\boldsymbol{\pi}}_a(s_t, \mathbf{u}_t)\\ 
&= Q^{\boldsymbol{\pi}}(s_t, \mathbf{u}_t) - \sum_{u' \in U}\pi^a(u'|\tau^a)Q^{\boldsymbol{\pi}}(s_t, \mathbf{u}_t^{-a}, u')\\ 
&\approx 
Q_\psi(s_t, \mathbf{u}_t) - \sum_{u' \in U}\pi^a(u'|\tau^a)Q_\psi(s_t, \mathbf{u}_t^{-a}, u'),
\end{align}
where $Q_\psi(s_t, \mathbf{u}_t^{-a}, u')$ is the action-value function evaluated by keeping the actions of all agents except $a$ fixed, and replacing agent $a$'s action with $u'$. The quantity: 
\begin{equation}
   \sum_{u' \in U}\pi^a(u'|\tau^a)Q^{\boldsymbol{\pi}}(s_t, \mathbf{u}_t^{-a}, u')
\end{equation}
is called the \textit{counterfactual} baseline.

In both CentralV and COMA, the critic $f_\psi$ is trained to minimise the loss $\mathcal{L}(\psi) = (y^{(\lambda)} - f_\psi(\cdot_t))^2$, using a variant of TD($\lambda$) \citep{sutton}. The targets $y^{(\lambda)} = (1 - \lambda)\sum_{n=1}^{\infty}\lambda^{n-1}G_t^{(n)}$ are a mixture of $n$-step returns: 
\begin{equation}
   G_t^{(n)} = \sum_{i=0}^{n-1}\gamma^{i}r_{t+i} + \gamma^n f_{\psi'}(\cdot_{t+n}),
\end{equation}
which are calculated using bootstrapped values from a target network \cite{Mnih2015}. Periodically, the target network parameters are copied from $\psi$.

\section{Method}
\label{sec:method}
In this section, we present our semi-on-policy (SOP) training framework and our COMA-CC algorithm.

\subsection{Semi-on-policy training}

The key insight behind our method is that, if the policy of the agents only changes slightly after a step of training, then the data generated by the previous policy can be reused in the next training step. In particular, we assume that if an episode $e$ was generated following a policy $\pi'$, and $\kl{\pi}{\pi'} \leq$ \texttt{kl\_threshold}, then we can still use the episode $e$ for training our policy $\pi$.\footnote{The choice of using the KL-divergence between policies to quantify similarity is somewhat arbitrary, and other criterions could also be used.}

Thus, if the base on-policy algorithm induces gradual policy updates, we can train it using the most recent episodes, despite the fact that some of them were generated by a different policy. To take advantage of this, we propose permissive semi-on-policy training (Algorithm \ref{algo:semi-on-policy}). Our results on SMAC show that this is an effective and reliable way of improving the sample efficiency of on-policy algorithms. For cases when the policy might diverge significantly during one update, we propose strict semi-on-policy training (Algorithm \ref{algo:semi-on-policy-alt}), which enforces the KL constraint between policies in the replay buffer. Because our analysis in Section \ref{sec:SOP_analysis} shows that permissive SOP training (henceforth simply called SOP training) is generally stable on SMAC, we defer evaluating strict SOP training to future work.

\begin{algorithm}[t!]
   \DontPrintSemicolon
   Let $b$ be the batch-size.\;
   Initialise the current policy $\pi$ and replay buffer \texttt{buf} of size $b$.\;
   Fill \texttt{buf} with $b$ episodes.\;
   \While{there is still time to train}{
      Train $\pi$ using all data in \texttt{buf}.*\;
      Discard the least recent episode from \texttt{buf}.\;
      Sample a new episode from the environment following the current policy $\pi$ and insert it into \texttt{buf}.\;
   }
   // (*) this step depends on how the particular policy gradient algorithm performs policy updates. For actor-critic algorithms: also train the critic.
   \caption{Permissive semi-on-policy training}
   \label{algo:semi-on-policy}
\end{algorithm}

\begin{algorithm}[t!]
   \DontPrintSemicolon
   Let $b$ be the batch-size.\;
   Let \texttt{kl\_threshold} be a hyperparameter.\;
   Initialise the current policy $\pi$, replay buffer \texttt{buf} of size $b$ and corresponding policy buffer \texttt{pbuf}. \;
   // Maintain the invariant that \texttt{buf[i]} was generated by following the policy \texttt{pbuf[i]}.\;
   \While{there is still time to train}{
      Sample new episodes from the environment following the current policy $\pi$ and insert them into \texttt{buf} until it is full. Fill corresponding locations of \texttt{pbuf} with $\pi$.\;
      Train $\pi$ using all data in \texttt{buf}.*\; 
      Discard the least recent episode from \texttt{buf} and all episodes generated by a sufficiently different policy.**\;
   }
   // (*) this step depends on how the particular policy gradient algorithm performs policy updates. For actor-critic algorithms: also train the critic.

   // (**) we say $\pi'$ is sufficiently different from $\pi$ if $\kl{\pi}{\pi'} >$ \texttt{kl\_treshold}.
   \caption{Strict semi-on-policy training}
   \label{algo:semi-on-policy-alt}
\end{algorithm}

\subsubsection{Implications on batch size}

Typical implementations of value-based MARL algorithms use a batch size of $32$ episodes on SMAC \citep{qmix_journal, wqmix, mahajan2019maven}. 
Because value-based methods generally use $Q$-learning \citep{qlearning} to train their value-function estimator \citep{qmix_journal, wqmix, mahajan2019maven}, they can reuse their old data. On-policy algorithms, however, must discard their data after training, so they typically opt for a smaller batch size \citep{qmix_journal}, a massive increase in the number of episodes sampled from the environment \citep{zhou2020learning}, or both \citep{papoudakis2020comparative}. If policy updates are stable across a small number of training steps, our method of training allows us to perform the same number of training steps as the currently popular off-policy value-based MARL methods, without requiring more data. 

\subsection{COMA-CC}

In this section we describe the improvements our COMA-CC algorithm makes to the original COMA.

\subsubsection{Fixing the critic inconsistency}
\label{ss:coma_critic_arch}
The COMA critic computes the counterfactual $Q$-function for each agent:
\begin{equation}
   Q^a(s_t, \mathbf{u}_t) := \seq{ Q(s_t, u_t^a=1, \mathbf{u}^{-a}_t), \dots, Q(s_t, u_t^a=|U|, \mathbf{u}^{-a}_t) }
   \label{eq:counter_qs}
\end{equation}
varying the action of agent $a$, while keeping other agents' actions fixed. In particular it estimates:
\begin{equation}
   Q^a(s_t, \mathbf{u}_t) \approx Q_{\psi}(s_t, z_t^{a}, \mathbf{u}_{t-1}, \mathbf{u}_t^{-a}, a)
\end{equation}
where $Q_{\psi}$ is a neural network with parameters $\psi$.
Here $s_t$ is the central state, $z_t^a$ is the observation of agent $a$ at time step $t$, $\mathbf{u}_{t-1}$ is the joint action at time step $(t-1)$, $\mathbf{u}_{t}^{-a}$ is the joint action at time step $t$ with the action of agent $a$ masked out, and $a$ is the agent id.

Consider a simple scenario where we have two agents, $a_1, a_2$, and they can perform a single action, $u$. Let $Q^{1} = f(s_t, z_t^1), Q^{2} = f(s_t, z_t^2)$ be the (neural network) estimates of the two agents' action-value functions. Since, in general, $z_t^1 \neq z_t^2$, this means that, in general, $Q^{1}(u, u) \neq Q^{2}(u, u)$. Thus, the agents' estimates for the \textit{actual} joint action taken at that time step are inconsistent with each other. 
The observation difference among agents is one high-level inconsistency that causes this issue, while another is having the agent id as input.
The way that $\mathbf{u}_t^{-a}$ is implemented (as a concatenation of one-hot action-vectors with one of the actions masked out) has a similar impact. 

One of the intuitive issues this causes is that two agents might disagree about whether the same joint action was good or bad. Thus, when they compute their respective share of the gradient, one will move the policy parameters towards making the joint action more likely, while the other will move away from it. Since the agents share parameters, this can cause convergence issues.

To address this issue, COMA-CC uses an alternative critic that estimates:
\begin{equation}
   Q(s_t, \mathbf{u}_t) \approx Q_{\phi}(s_t, z_t^{1}, \dots, z_t^{{n}}, \mathbf{u}_{t-1}, \mathbf{u}_t)
\end{equation}
We can use this critic to compute each component of $Q^a(s_t, \mathbf{u}_t)$ by inputting the respective counterfactual joint actions.
Furthermore, this critic always receives the same inputs when estimating the same joint action, resolving the issue of inconsistent estimates.

The new critic can compute all necessary counterfactual $Q$-values in a single forward pass; this is done by building the input for each counterfactual $Q$-value computation and running the concatenated inputs through the network. For each counterfactual $Q$-value computation, the COMA critic requires $n$ inputs, one for each agent, and the   COMA-CC critic requires $nm$ inputs, one for each agent and counterfactual joint action; therefore, if the internal network architectures are similar for both critics, the   COMA-CC critic must perform $O(m)$ times more multiplications than the COMA critic. To ensure this, the concatenated observations $(z_t^1, \dots, z_t^{n})$ must be compressed before they are used as inputs to the critic; this can be done with an encoding network. Since the running time of our experiments was essentially the same for both the COMA and   COMA-CC critics, we do not include this optimisation in our implementation.\footnote{A more thorough analysis of the computational complexity of the COMA-CC critic is included in Appendix \ref{appx:coma_alt_critic}. }

\subsubsection{Critic training}
\label{sec:critic_training}

The critic in COMA is trained according to the procedure given in Algorithm \ref{algo:coma_critic_train}.
This way of training means that each training iteration uses only \texttt{batch\_size} steps of agent-environment interaction to compute the gradients: for example, if the \texttt{batch\_size} is $8$ episodes, and each episode has length $128$, COMA will train on minibatches of $8$ steps until it trains on all data from the episodes. Our ablative study in Section \ref{sec:coma_ablations} indicates that this destabilises learning; therefore, we train COMA-CC on the whole batch at once. The change necessary is to the for-loop, and is given in Algorithm \ref{algo:coma_critic_train_alt}.

\begin{algorithm}[t]
   \DontPrintSemicolon
   Initialise the critic parameters $\phi$, target critic parameters $\hat{\phi}$ = $\phi$, and policy parameters $\theta$\;
   \dots\;
   [Let $T$ be the maximum length of an episode.]\;
   // For all agents $a \in A$ simultaneously:\;
   Calculate TD($\lambda$) targets $y_t^a$ using the target critic $\hat{\phi}$\;
   \For{t = T \KwTo 1}{
      $\Delta Q_t^a = y_t^a - Q_{\phi}(s_t, \mathbf{u}_t)$\;
      $\Delta \phi = \nabla_{\phi}(\Delta Q_t^a)^2$ // calculate critic gradient\;
      $\phi = \phi - \alpha \Delta \phi$ // update critic parameters\;
      Every $C$ steps reset $\hat{\phi} = \phi$\;
   }
   \dots\;
   \caption{COMA Critic Training \citep{coma}}
   \label{algo:coma_critic_train}
\end{algorithm}

\begin{algorithm}[t]
   \DontPrintSemicolon
   $\Delta \phi = 0$\;
   \For{t = T \KwTo 1}{
      $\Delta Q_t^a = y_t^a - Q_{\phi}(s_t, \mathbf{u}_t)$\;
      $\Delta \phi = \Delta \phi + \nabla_{\phi}(\Delta Q_t^a)^2$ // accumulate critic gradient\;
   }
   $\phi = \phi - \alpha \Delta \phi$ // update critic parameters\;
   Every $C$ steps reset $\hat{\phi} = \phi$\;
   \caption{Alternative COMA Critic Training}
   \label{algo:coma_critic_train_alt}
\end{algorithm}

\section{Experimental setup}
\label{sec:experimental_setup}
\begin{figure*}[t!]
	\centering
   \begin{subfigure}[b]{0.3\textwidth}
       \includegraphics[width=\textwidth]{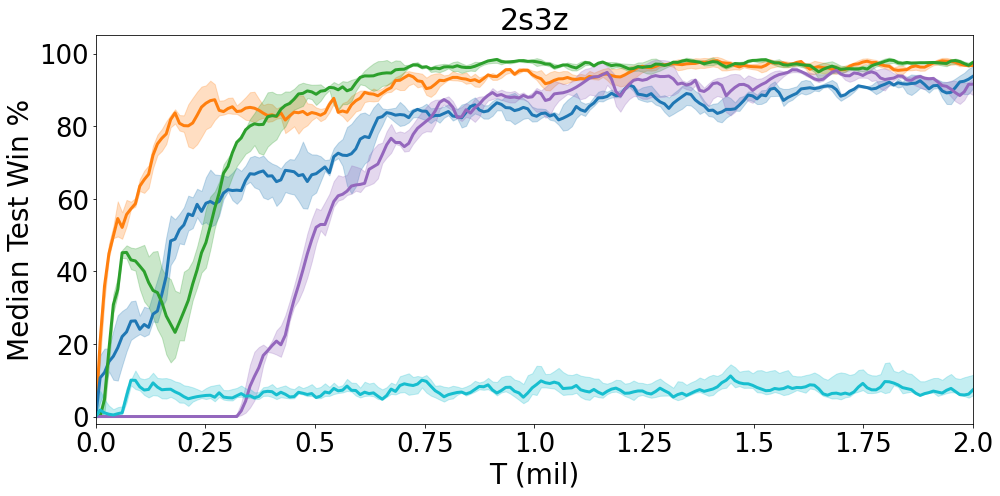}
   \end{subfigure}
   \hfill
   \begin{subfigure}[b]{0.3\textwidth}
       \includegraphics[width=\textwidth]{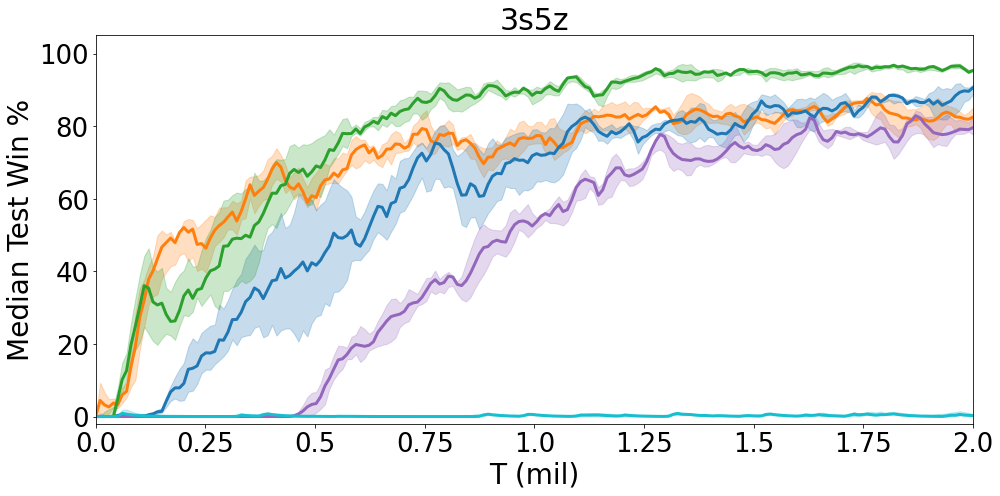}
   \end{subfigure}
   \hfill
   \begin{subfigure}[b]{0.3\textwidth}
       \includegraphics[width=\textwidth]{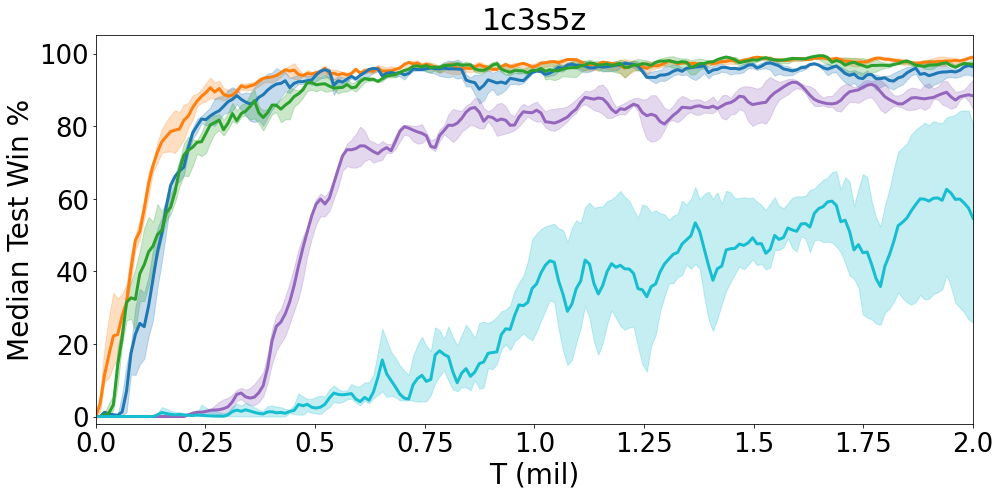}
   \end{subfigure}

   \begin{subfigure}[b]{0.3\textwidth}
      \centering
      \includegraphics[width=\textwidth]{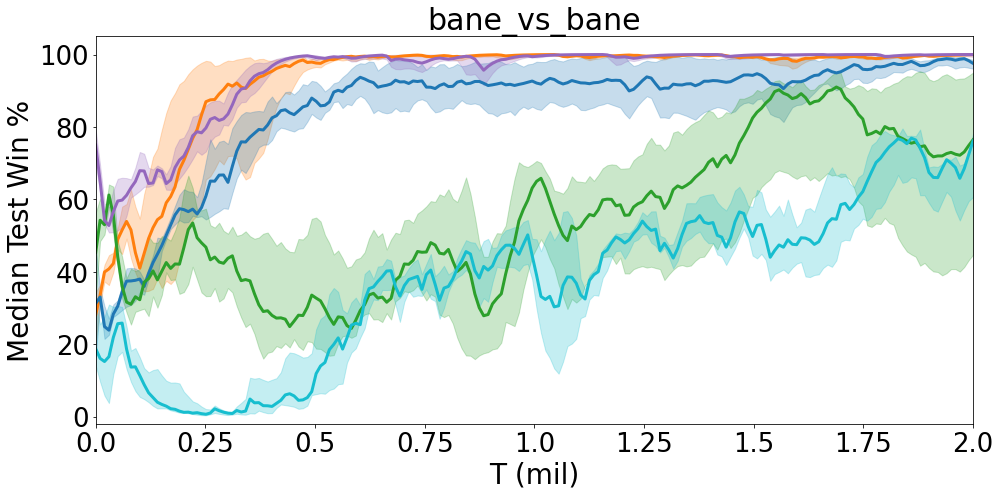}
  \end{subfigure}
  \hfill
  \begin{subfigure}[b]{0.3\textwidth}
      \centering
      \includegraphics[width=\textwidth]{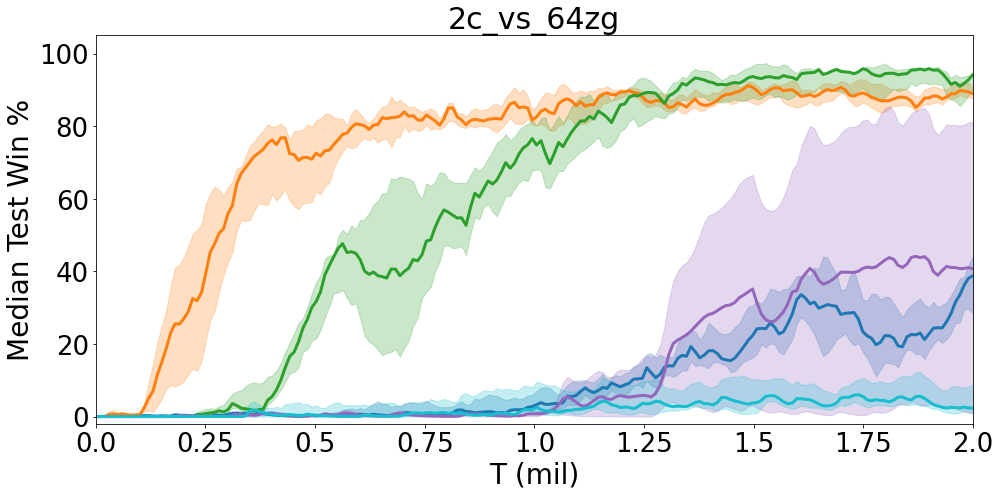}
  \end{subfigure}
  \hfill
  \begin{subfigure}[b]{0.3\textwidth}
      \centering
      \includegraphics[width=\textwidth]{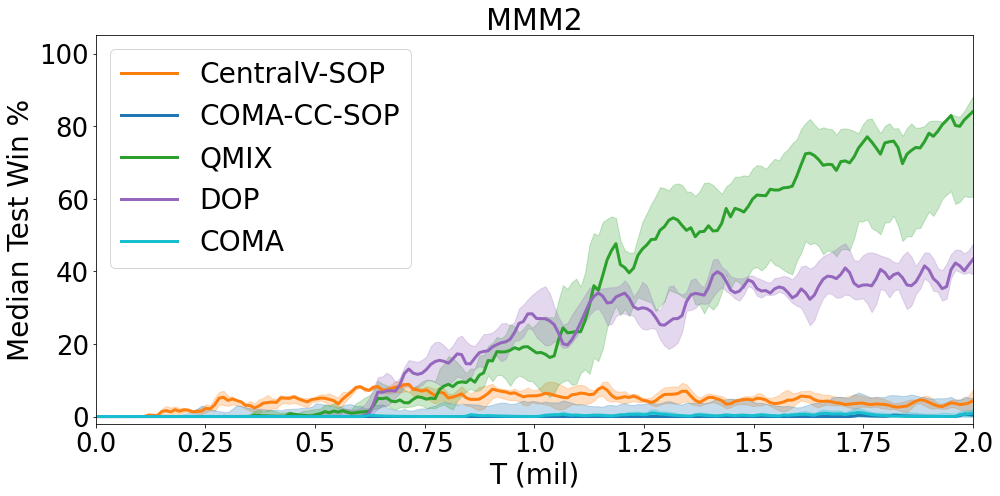}
  \end{subfigure}
 
   \caption{Comparing MAPG methods with QMIX. Batch size $32$. Easy maps (top): \texttt{[2s3z, 3s5z, 1c3s5z]}, Hard maps (bottom): \texttt{[bane\_vs\_bane, 2c\_vs\_64zg]}, Super-hard maps (bottom): \texttt{[MMM2]}.}
   \label{fig:ac_vs_qmix}
   \Description{6 plots with titles 2s3z, 3s5z, 1c3s5z, bane\_vs\_bane, 2c\_vs\_64zg, MMM2. X-axis labelled T (mil). Y-axis labelled Median Test Win \%. Legend contains CentralV-SOP,   COMA-CC-SOP, QMIX, DOP.}
\end{figure*}

We evaluate all methods on the challenging SMAC benchmark \citep{smac}, using a batch size of either $8$ or $32$ episodes. In all comparisons, all methods use the same batch size; we indicate this in the figure notes. We run $4$ random seeds per map for each method. 

For QMIX and DOP we use the author-provided open-source implementations.\footnote{\url{https://github.com/oxwhirl/pymarl} and \url{https://github.com/TonghanWang/DOP}.} We create fresh implementations of CentralV and COMA-CC, based on the existing CentralV and COMA implementations.\footnote{\url{https://github.com/oxwhirl/pymarl}.} The changes we make are outlined in Section \ref{sec:method}. Additionally, we set $\gamma = 1$ when computing the advantage function for CentralV, which we find helps stabilise policy updates; we include results for the version without this change in Appendix \ref{appx:centralv_default}.

All experiments, including baselines, use StarCraft II Version \texttt{2.4.10}.\footnote{At the time of writing, the latest version of the PyMARL framework provided by the original SMAC authors uses \texttt{SC2.4.10}. Older versions of the framework used \texttt{SC2.4.6.2.69232}. The SMAC authors note that results are not always comparable between versions. We have retested all baselines to ensure fair comparisons.} We use the same hyperparameters, network architectures, and evaluation methodology as in the original SMAC paper \cite{smac}, unless stated otherwise. We list these in detail in Appendix \ref{appx:setup}. Doing so, we have attempted to eliminate all possible confounding factors in our COMA-CC ablation study and in our comparisons of CentralV and COMA-CC. As such, our results should be interpreted as purely comparative, and we believe that better absolute results are achievable in practice, with proper tuning.

We focus on $6$ maps in this section: $3$ easy maps \verb|2s3z|, \verb|3s5z|, \verb|1c3s5z|, $2$ hard maps \verb|bane_vs_bane|, \verb|2c_vs_64zg|, and $1$ super-hard map \verb|MMM2|. The easy maps were chosen because of the surprisingly low performance of CentralV and COMA reported previously in the literature \cite{smac,papoudakis2020comparative,qmix_journal,wang2020dop}.\footnote{Some of these papers use the older version of StarCraft II. However, our experiments indicate that the low performance is not due to version number alone.}  
The hard maps include a challenging symmetric and asymmetric scenario, and the super-hard map tests the limits of our method.
For each map, we report the median test win-rate across all seeds, and we shade the area between the 25\textsuperscript{th} and 75\textsuperscript{th} percentile. For aggregate results, we plot the averaged median win-rate across all seeds and all $14$ SMAC maps for the given algorithm. Finally, when we refer to the entire SMAC challenge, we mean the $14$ maps explicitly mentioned in the original paper \cite{smac}; in particular, we do not consider the additional maps distributed with the package. The algorithms that use semi-on-policy training are suffixed with "-SOP".

\begin{figure}
   \centering
   \includegraphics[width=0.8\columnwidth]{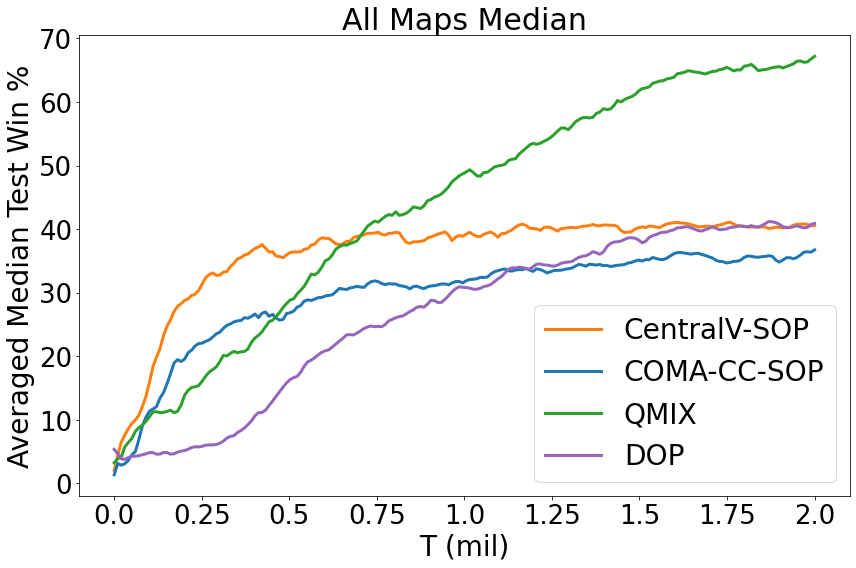}
   \caption{Averaged median test win rate for policy gradient algorithms and QMIX across all 14 SMAC scenarios.
}
   \label{fig:ac_vs_qmix_all_maps}
   \Description{1 plot with title All Maps Median. X-axis labelled T (mil). Y-axis labelled Averaged Median Test Win \%. Legend contains CentralV-SOP,   COMA-CC-SOP, QMIX, DOP.}
\end{figure}

\begin{figure*}[t!]
	\centering
   \begin{subfigure}[b]{0.3\textwidth}
       \includegraphics[width=\textwidth]{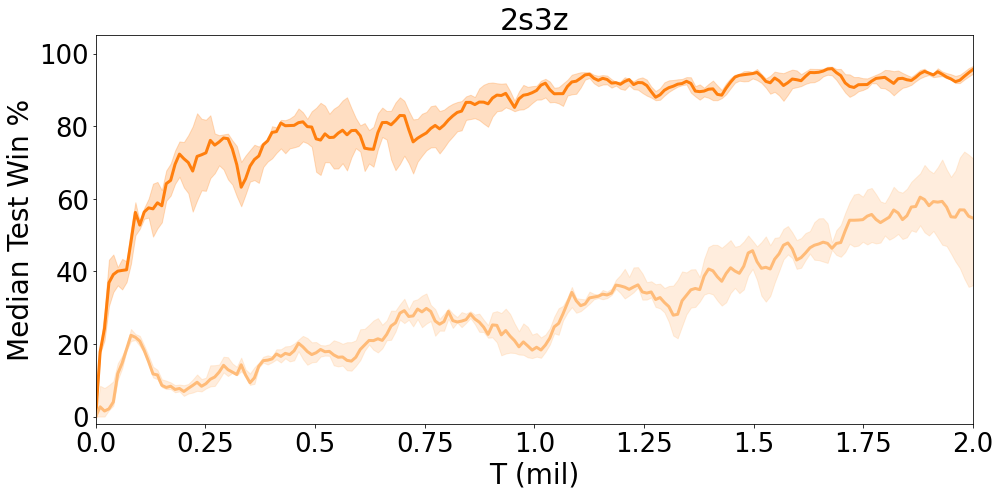}
   \end{subfigure}
   \hfill
   \begin{subfigure}[b]{0.3\textwidth}
       \includegraphics[width=\textwidth]{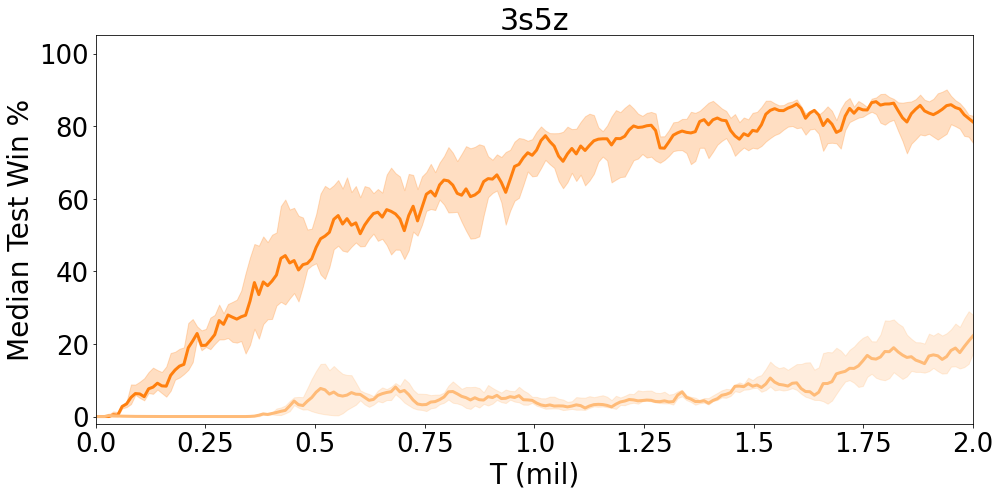}
   \end{subfigure}
   \hfill
   \begin{subfigure}[b]{0.3\textwidth}
       \includegraphics[width=\textwidth]{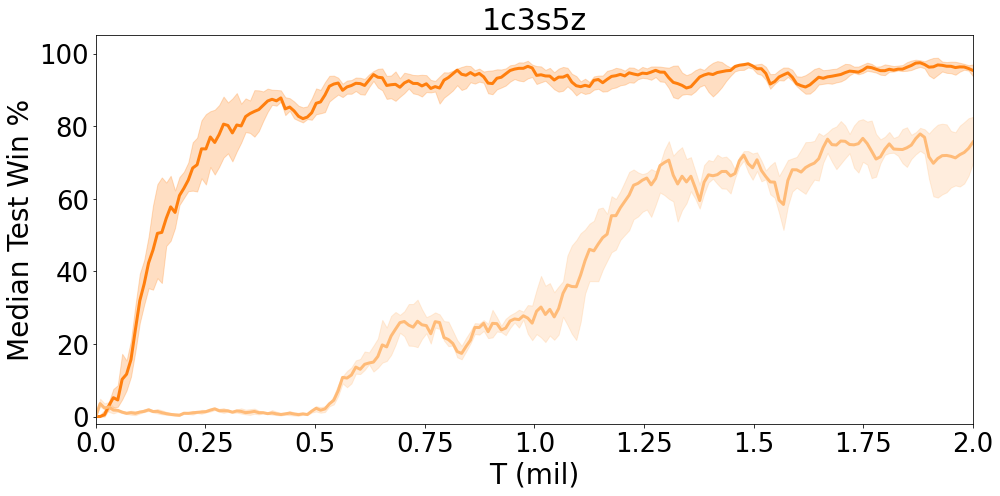}
   \end{subfigure}

   \begin{subfigure}[b]{0.3\textwidth}
      \centering
      \includegraphics[width=\textwidth]{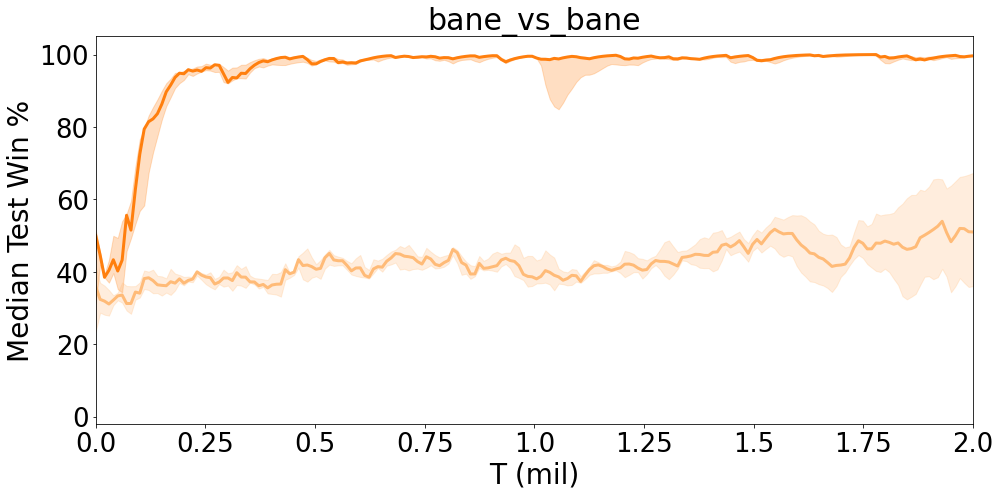}
  \end{subfigure}
  \hfill
  \begin{subfigure}[b]{0.3\textwidth}
      \centering
      \includegraphics[width=\textwidth]{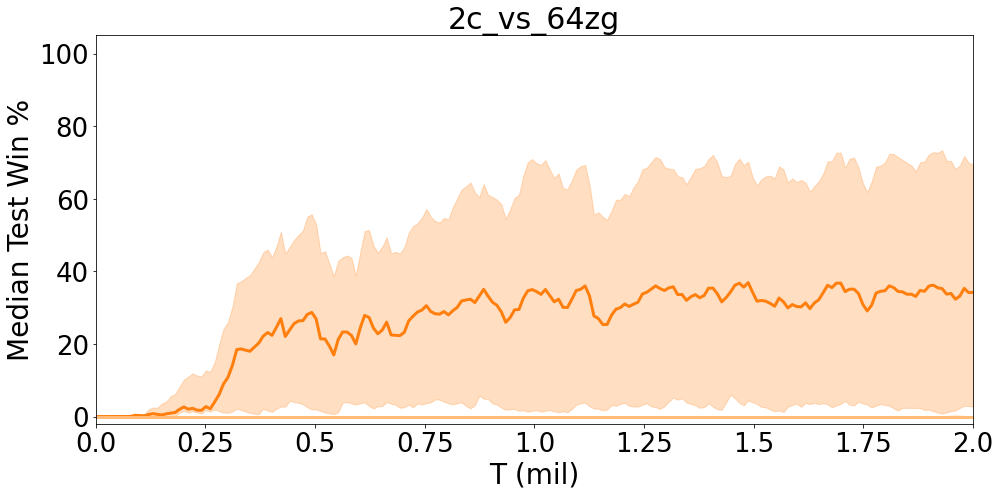}
  \end{subfigure}
  \hfill
  \begin{subfigure}[b]{0.3\textwidth}
      \centering
      \includegraphics[width=\textwidth]{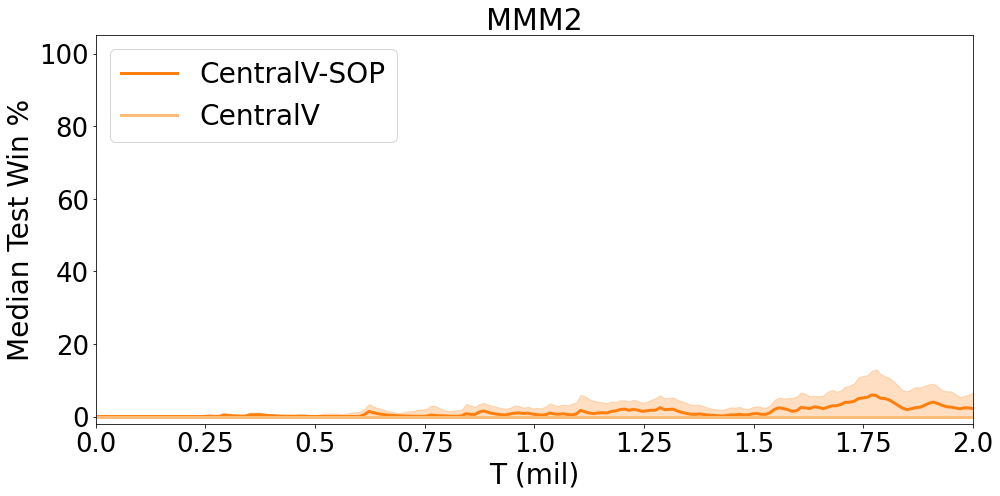}
  \end{subfigure}
 
   \caption{Comparing CentralV with SOP training and with purely on-policy training. Batch size $8$.}
   \label{fig:centralv_sop_vs_op}
   \Description{6 plots with titles 2s3z, 3s5z, 1c3s5z, bane\_vs\_bane, 2c\_vs\_64zg, MMM2. X-axis labelled T (mil). Y-axis labelled Median Test Win \%. Legend contains CentralV-SOP and CentralV.}
\end{figure*}

\section{Results}
\label{sec:results}
\begin{figure*}[t!]
	\centering
   \begin{subfigure}[b]{0.3\textwidth}
       \includegraphics[width=\textwidth]{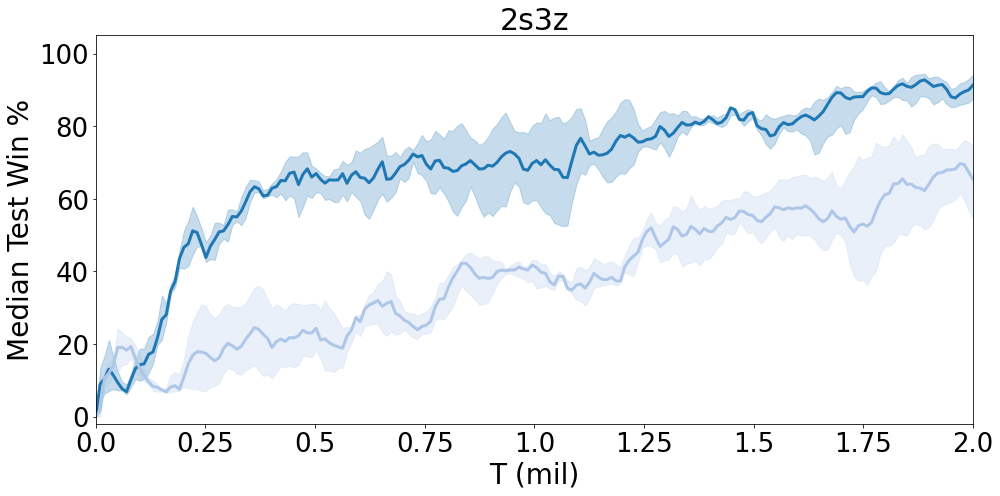}
   \end{subfigure}
   \hfill
   \begin{subfigure}[b]{0.3\textwidth}
       \includegraphics[width=\textwidth]{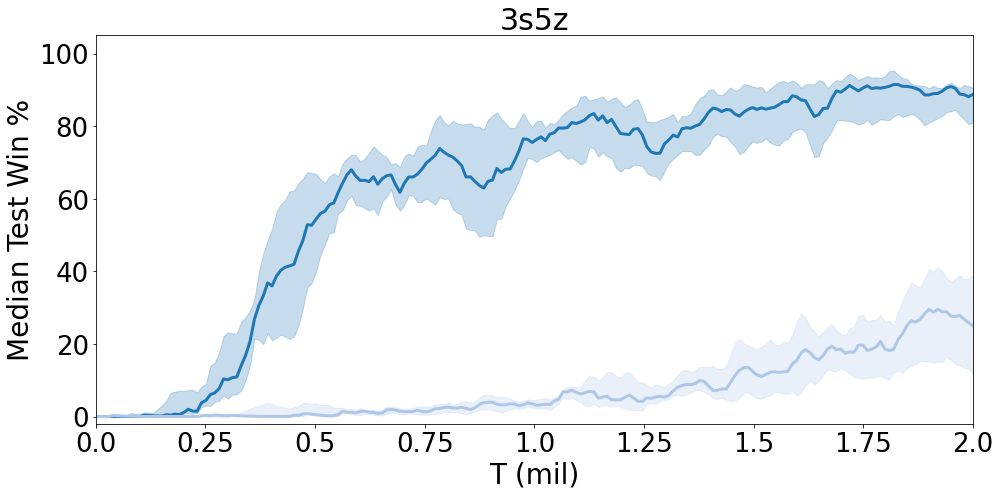}
   \end{subfigure}
   \hfill
   \begin{subfigure}[b]{0.3\textwidth}
       \includegraphics[width=\textwidth]{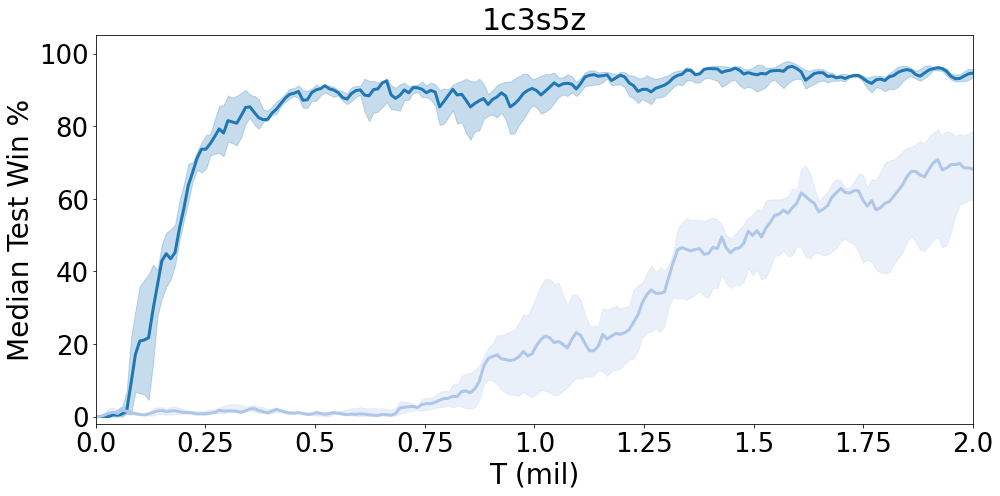}
   \end{subfigure}

   \begin{subfigure}[b]{0.3\textwidth}
      \centering
      \includegraphics[width=\textwidth]{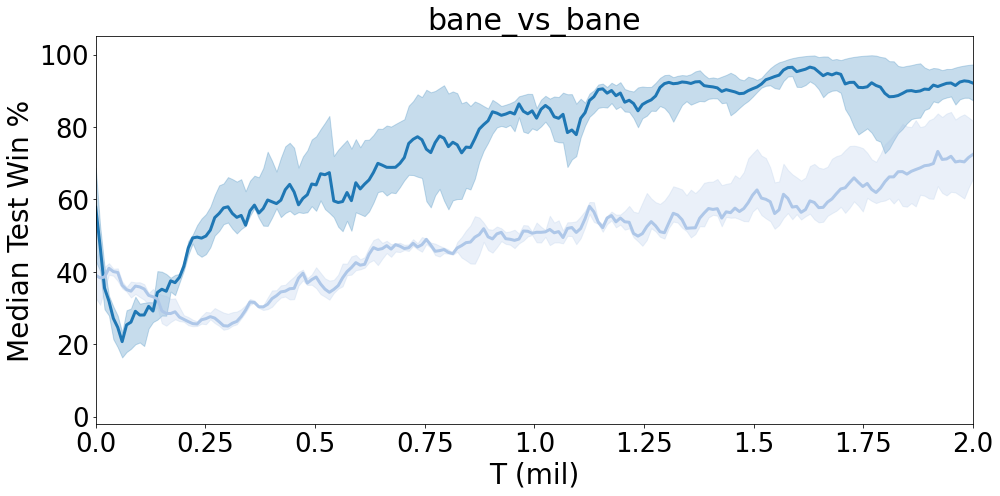}
  \end{subfigure}
  \hfill
  \begin{subfigure}[b]{0.3\textwidth}
      \centering
      \includegraphics[width=\textwidth]{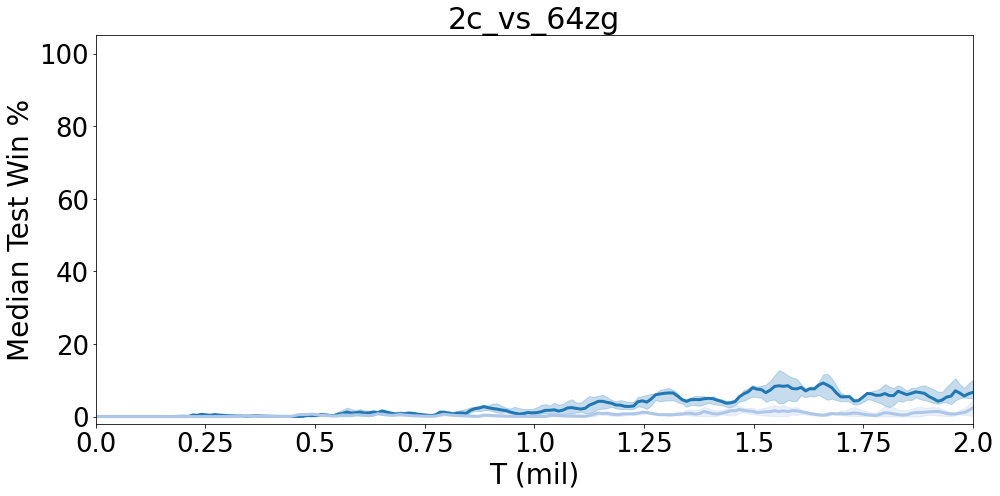}
  \end{subfigure}
  \hfill
  \begin{subfigure}[b]{0.3\textwidth}
      \centering
      \includegraphics[width=\textwidth]{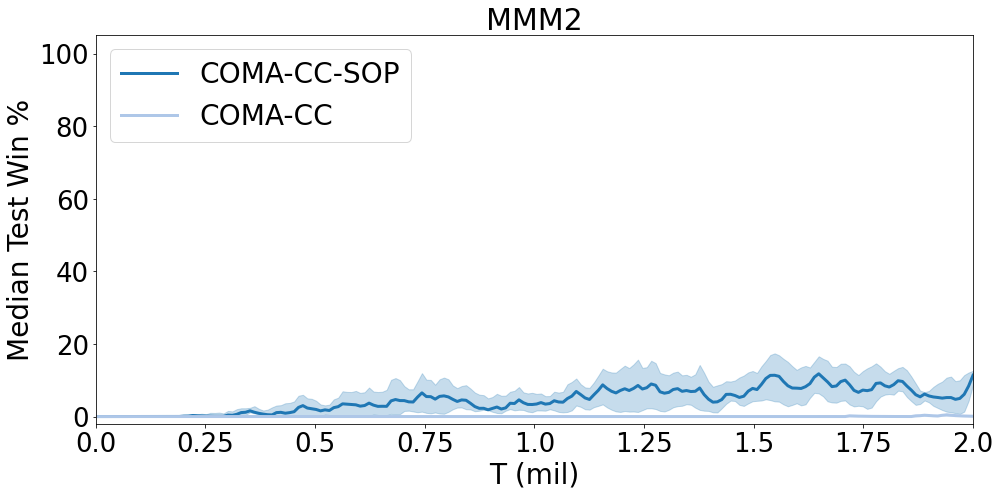}
  \end{subfigure}
 
   \caption{Comparing COMA-CC with SOP training and with purely on-policy training. Batch size $8$.}
   \label{fig:coma_sop_vs_op}
   \Description{6 plots with titles 2s3z, 3s5z, 1c3s5z, bane\_vs\_bane, 2c\_vs\_64zg, MMM2. X-axis labelled T (mil). Y-axis labelled Median Test Win \%. Legend contains   COMA-CC-SOP and   COMA-CC.}
\end{figure*}

Figure \ref{fig:ac_vs_qmix} shows that CentralV-SOP and COMA-CC-SOP perform on par with the state-of-the-art value-based method QMIX on various SMAC maps, and even outperform it on some maps. In particular, QMIX fails to converge on \verb|bane_vs_bane|, whereas all MAPG methods manage to solve the scenario. Additionally, CentralV-SOP achieves the same performance as QMIX on \verb|2c_vs_64zg|, outperforming all other MAPG methods and converging much faster.  
The best performing MAPG method is CentralV-SOP, which outperforms other MAPG methods on most maps. Furthermore, COMA-CC-SOP significantly outperforms the original COMA. Results for all 14 SMAC maps are included in Appendix \ref{appx:full_results}.

Figure \ref{fig:ac_vs_qmix_all_maps} shows the aggregate results on the entire SMAC challenge. QMIX performs the best on average. However, as previously discussed, it is not the best on all maps. CentralV-SOP and DOP achieve approximately the same performance at the end of training, with CentralV-SOP converging much faster. One significant difference between CentralV and COMA or DOP is that the former learns a centralised $V$-function, while the latter learn a centralised $Q$-function, which could be more difficult to learn in the multi-agent setting due to the exponentially sized joint-action space.

Figures \ref{fig:centralv_sop_vs_op} and \ref{fig:coma_sop_vs_op} show the results comparing SOP training to on-policy training for CentralV and COMA-CC respectively. 
We see that the algorithms trained with SOP data consistently outperform the corresponding on-policy versions, demonstrating better sample efficiency and stable learning.

\section{Analysis}
\label{sec:analysis}
\subsection{SOP Training}
\label{sec:SOP_analysis}
\begin{figure*}
   \begin{subfigure}[b]{0.3\textwidth}
       \includegraphics[width=\textwidth]{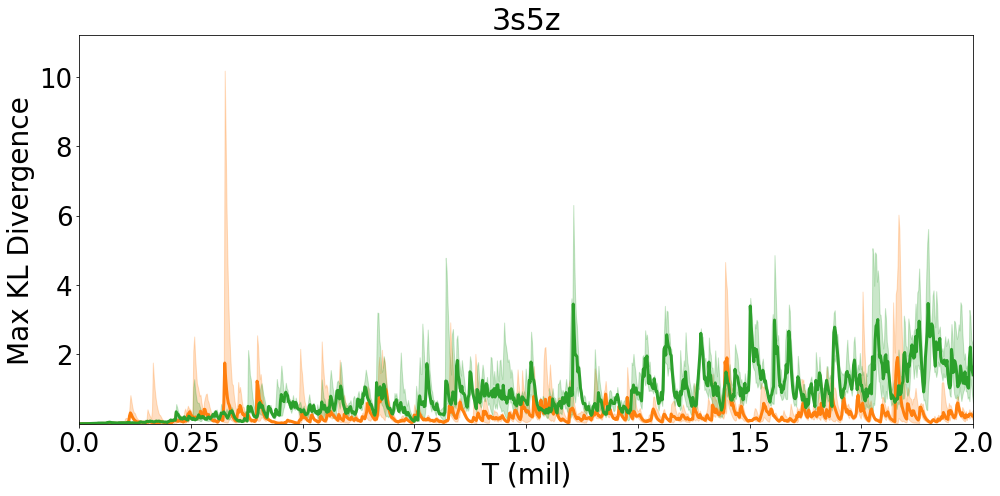}
   \end{subfigure}
   \hfill
   \begin{subfigure}[b]{0.3\textwidth}
       \includegraphics[width=\textwidth]{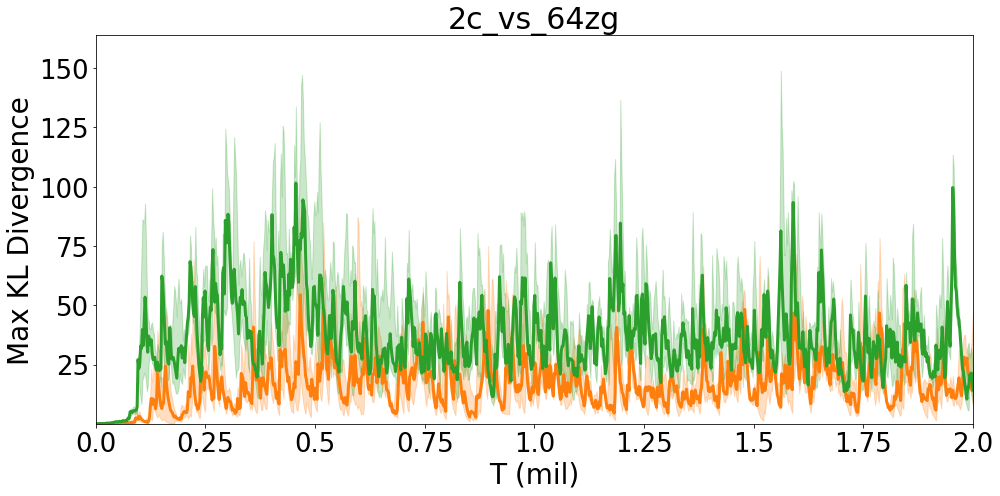}
   \end{subfigure}
   \hfill
   \begin{subfigure}[b]{0.3\textwidth}
       \includegraphics[width=\textwidth]{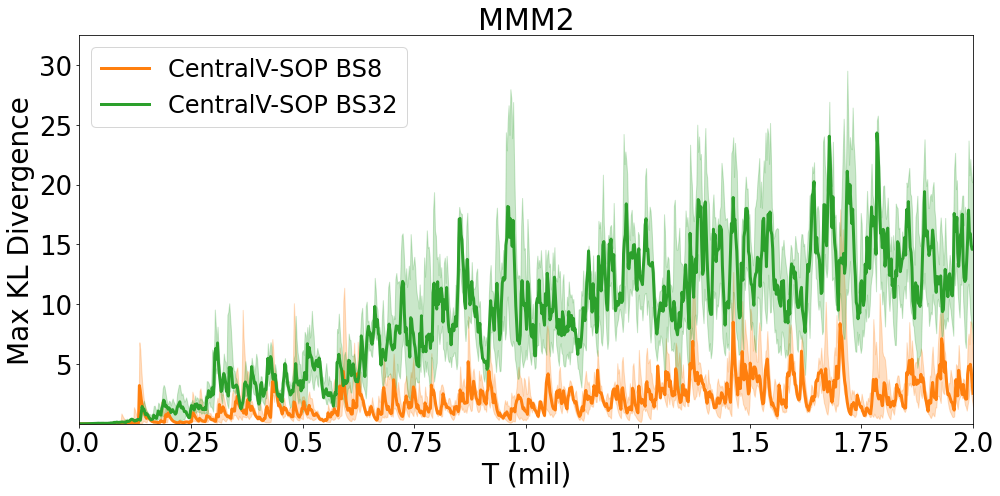}
   \end{subfigure}
   \caption{Maximum estimated KL divergence between the current policy $\pi$ and any other policy $\pi'$ in the buffer. Median over all seeds. BS indicates batch size.}
   \label{fig:sop_kl_analysis}
   \Description{3 plots with titles 3s5z, 2c\_vs\_64zg, MMM2. X-axis labelled T (mil). Y-axis labelled Max KL Divergence. Legend contains CentralV-SOP BS8 and CentralV-SOP BS32.}
\end{figure*}

In this section we provide more insight into when SOP training works and how to best utilise it. Figure \ref{fig:sop_kl_analysis} shows a plot of the maximum KL divergence between the current policy $\pi$ and any other policy in the buffer for various maps. To estimate the KL divergence, we use an unbiased estimator \cite{sch_kl}:
\begin{equation}
   \kl{p}{q} = \mathbb{E}_{x \sim p}\left[\frac{q(x)}{p(x)} - 1 - \log \frac{q(x)}{p(x)}\right]
\end{equation}
As before, we plot the median across seeds and shade the area between the 25\textsuperscript{th} and 75\textsuperscript{th} percentile. In the map \texttt{3s5z} we can see that the divergence is extremely low: less than $4$ in most cases, and going up to at most $10$ for the 75\textsuperscript{th} percentile. This fits our intuitive description that we can use past data, as long as the current policy has not diverged too much from the policy that generated it. However, even divergences as high as $150$ can still yield stable training, as the results for \texttt{2c\_vs\_64zg} indicate. The reason is that the action space in \texttt{2c\_vs\_64zg} is much larger than most other maps, and there are many actions with a similar effect; thus, while the KL divergence may rise, the underlying behaviour remains approximately the same.

In practice we find that sudden big changes to the current policy lead to instability when using SOP training. There are different ways to remedy this: for example modifying critic training or advantage estimation, lowering the learning rate, or using strict SOP training (Algorithm \ref{algo:semi-on-policy-alt}) with an appropriate \texttt{kl\_threshold}. We hypothesise that, in most cases, permissive SOP training (Algorithm \ref{algo:semi-on-policy}) would be preferable, since, if the policy changes are unstable, strict SOP training would prune too much of the old data and essentially degenerate to on-policy training. The KL estimator is also an important factor to consider when determining appropriate values of \texttt{kl\_threshold}. Additionally, our results from Figure \ref{fig:sop_kl_analysis} indicate that, if used as a hyperparameter, it should be tuned on a per-application basis (i.e., per map in the case of SMAC). However, our results from Section \ref{sec:results} indicate that permissive SOP training suffices for SMAC.

\subsection{COMA Ablations}
\label{sec:coma_ablations}

\begin{figure*}
   \centering
   \begin{subfigure}[t!]{0.6\textwidth}
       \centering
       \includegraphics[width=\textwidth]{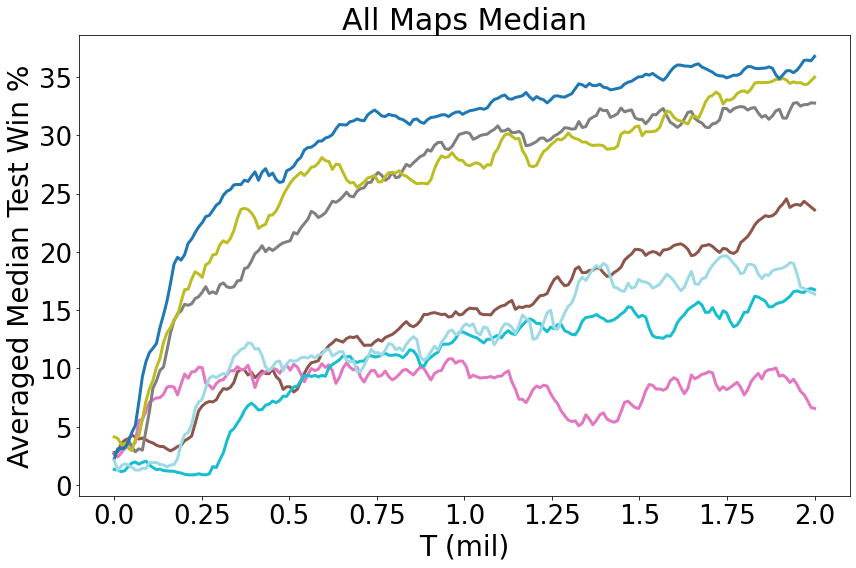}
   \end{subfigure}
   \hfill
   \begin{subfigure}[t!]{0.3\textwidth}
       \centering
       \includegraphics[width=\textwidth]{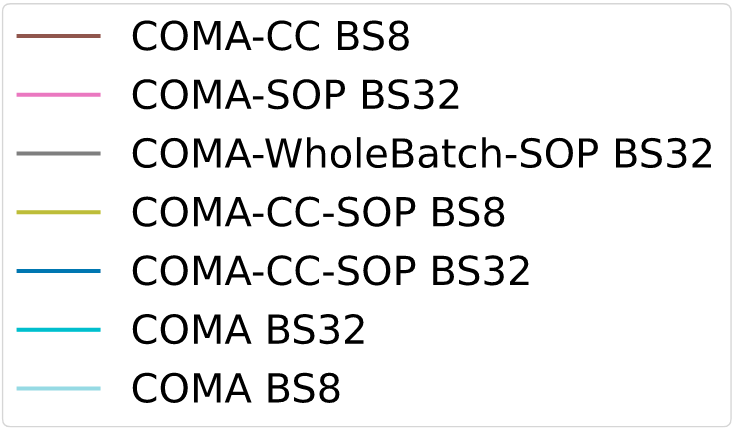}
   \end{subfigure}
   \caption{Averaged median test win rate for   COMA-CC ablations across all 14 SMAC scenarios. BS indicates batch size.}
   \label{fig:coma_ablations}
   \Description{1 plot with title All Maps Median. X-axis labelled T (mil). Y-axis labelled Averaged Median Test Win \%. Legend contains   COMA-CC BS8, COMA-SOP BS32, COMA-WholeBatch-SOP BS32,   COMA-CC-SOP BS8,   COMA-CC-SOP BS8,   COMA-CC-SOP BS32, COMA BS32, COMA BS8.}
\end{figure*}

We now analyse the effect of each change we made to the COMA algorithm. Figure \ref{fig:coma_ablations} shows the averaged median test win rate across all $14$ SMAC scenarios. We use the same naming convention as before, with the addition of the \texttt{COMA-WholeBatch} algorithm, which incorporates new training from Section \ref{sec:critic_training}, but does not use the new critic proposed in Section \ref{ss:coma_critic_arch}.

We find that COMA-CC-SOP outperforms all ablations. Training on the entire batch at once and using SOP training are crucial to achieving the best results with COMA. We find that it is important to have both additions, as \texttt{COMA-SOP} (just doing SOP training) and \texttt{COMA-CC} (just doing whole-batch training), produce far weaker results compared to the ablations that incorporate both changes.  
Fixing the critic representation also yields some improvements, though they are less significant than we expected, as \texttt{COMA-WholeBatch-SOP} performs similarly to \texttt{COMA-CC-SOP}. Additionally, we do not find significant differences between using a batch size of $32$ episodes, compared to using a batch size of $8$ episodes. Overall, COMA-CC-SOP achieves a more than $200\%$ higher average median test win-rate compared to the original COMA on the SMAC benchmark. Additionally, it exhibits more stable learning.

\section{Conclusion}

In this paper we proposed a simple to implement enhancement that yielded significant improvements to the performance and sample efficiency of both multi-agent policy gradient algorithms we examined: namely, using semi-on-policy training. With this enhancement, we established CentralV-SOP as a strong benchmark for multi-agent policy gradient algorithms on SMAC, showing that it is competitive with state-of-the-art value-based and policy gradient methods. The fact that CentralV-SOP often outperforms COMA and DOP, two methods that address the multi-agent credit assignment problem, indicates that credit assignment might be an insignificant factor on SMAC.

\balance
\bibliographystyle{ACM-Reference-Format}
\bibliography{bibliography}

\clearpage
\appendix
\section{Experimental setup}
\balance
\label{appx:setup}
All environment configurations for SMAC are kept at their default values. We use the same network architecture and hyperparameters as in \cite{smac}, unless stated otherwise. The configuration is the same for both CentralV and COMA-CC. We present it here for completeness.

\subsection{Network architecture}

The policy network is a recurrent network which contains a GRU of hidden dimension $64$ with a fully-connected linear layer before and after. To speed up learning, parameters are shared across agents. The critic architecture for both algorithms is a fully-connected feedforward neural network, with $2$ hidden layers of $128$ units each, and a final layer of $1$ unit; as discussed in Section \ref{ss:coma_critic_arch}, the original COMA critic has a final layer of $|U|$ units. On each training iteration, we first update the critic, and then the actor, performing a single gradient descent step in both cases; the original implementations of these algorithms used minibatch gradient descent to update their critics, as in Algorithm \ref{algo:coma_critic_train} from Section \ref{sec:critic_training}. Both networks apply a ReLU non-linearity between all but the final layer.

\subsection{Hyperparameters}
For both CentralV and COMA-CC (including ablations) we use the following hyperparameters: for TD($\lambda$) we set $\lambda = 0.8$ for all maps, with discount factor $\gamma = 0.99$. Both algorithms update the target critic every $200$ training iterations. The actor and critic are optimised separately using an RMSProp optimiser; in both cases we set the optimiser parameters as follows: learning rate $ = 0.005$, $\alpha = 0.99$, and $\varepsilon = 1$e-$5$. To enforce exploration, both algorithms use an $\epsilon$-floor scheme (as in \cite{coma}) for the agents' policies, with $\epsilon$ linearly annealed from $0.5$ to $0.01$ over 100k time steps. As in \cite{coma, smac}, at test time, the algorithms perform greedy action selection, choosing the actions with highest probability.

\section{Computational complexity of the   COMA-CC critic}
\label{appx:coma_alt_critic}
Multiplying a $n \times k$ matrix with a $k \times m$ matrix requires $O(n \times k \times m)$ multiplications. The COMA critic is a neural network, which takes $i_1$ inputs, has two hidden layers of size $h$, and produces $m$ outputs. The   COMA-CC critic takes $i_2$ inputs, has two hidden layers of size $h$, and produces a single output. 
Thus, on a single input, the COMA critic requires $O(hi_1) + O(h^2) + O(hm)$ multiplications to compute its outputs and the   COMA-CC critic requires $O(hi_2) + O(h^2) + O(h)$ multiplications. 

Note that $h$ is typically a fixed constant. Furthermore, note that $m = O(\min \set{i_1,i_2})$, since the joint action is included in the input. Thus, we get that the COMA critic requires $O(i_1)$ multiplications and the   COMA-CC critic requires $O(i_2)$ multiplications. For each baseline computation, the COMA critic takes $n$ inputs, one for each agent, so requires $O(n \times i_1)$ multiplications; the COMA-CC critic takes $nm$ inputs, one for each agent and counterfactual joint action, so requires $O(n \times m \times i_2)$ multiplications. Therefore, assuming $i_1 = \Theta(i_2)$, the COMA-CC critic network performs $O(m)$ times as many multiplications as the COMA critic.

\section{Full results}
\label{appx:full_results}
In this section, we present our full set of results.

\subsection{Actor-critic algorithms vs QMIX}

\begin{figure*}
   \centering
   \begin{subfigure}[b]{0.3\textwidth}
       \centering
       \includegraphics[width=\textwidth]{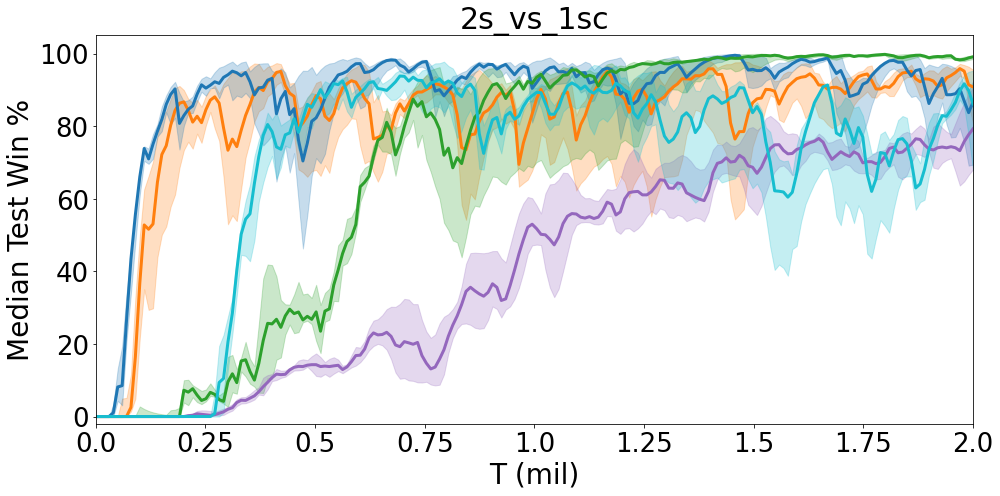}
   \end{subfigure}
   \hfill
   \begin{subfigure}[b]{0.3\textwidth}
       \centering
       \includegraphics[width=\textwidth]{ac_vs_qmix/2s3z}
   \end{subfigure}
   \hfill
   \begin{subfigure}[b]{0.3\textwidth}
       \centering
       \includegraphics[width=\textwidth]{ac_vs_qmix/3s5z}
   \end{subfigure}

   \begin{subfigure}[b]{0.3\textwidth}
     \centering
     \includegraphics[width=\textwidth]{ac_vs_qmix/1c3s5z}
   \end{subfigure}
   \hfill
   \begin{subfigure}[b]{0.3\textwidth}
      \centering
      \includegraphics[width=\textwidth]{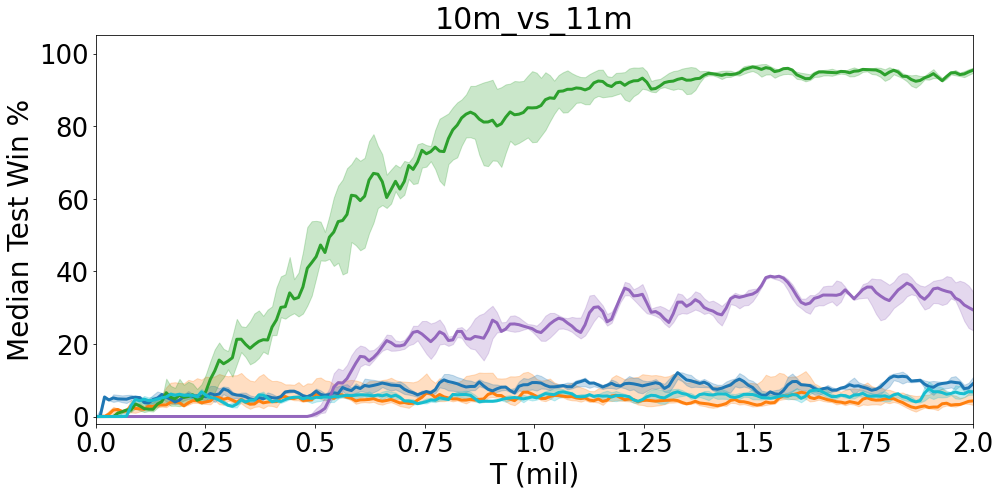}
   \end{subfigure}
   \hfill
   \begin{subfigure}[b]{0.3\textwidth}
      \centering
      \includegraphics[width=\textwidth]{ac_vs_qmix/2c_vs_64zg}
   \end{subfigure}

   \begin{subfigure}[b]{0.3\textwidth}
      \centering
      \includegraphics[width=\textwidth]{ac_vs_qmix/bane_vs_bane}
   \end{subfigure}
   \hfill
   \begin{subfigure}[b]{0.3\textwidth}
      \centering
      \includegraphics[width=\textwidth]{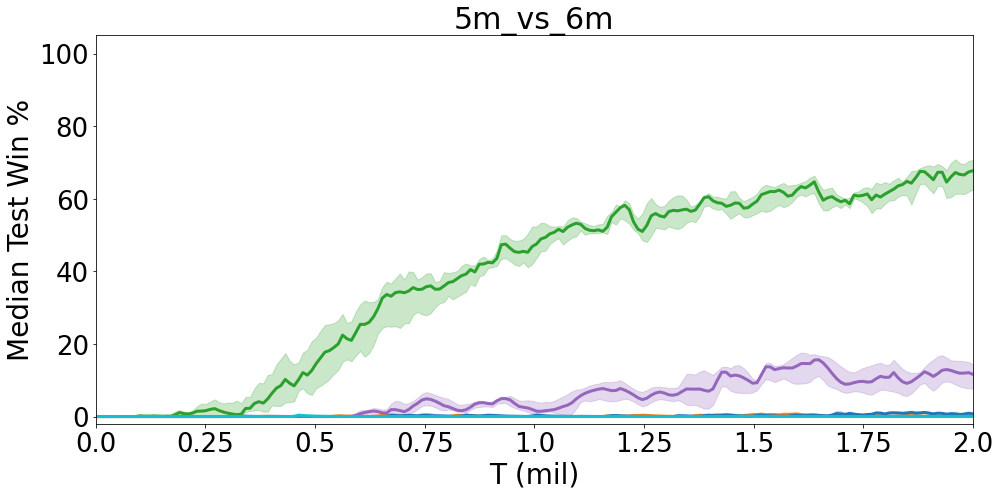}
   \end{subfigure}
   \hfill
   \begin{subfigure}[b]{0.3\textwidth}
      \centering
      \includegraphics[width=\textwidth]{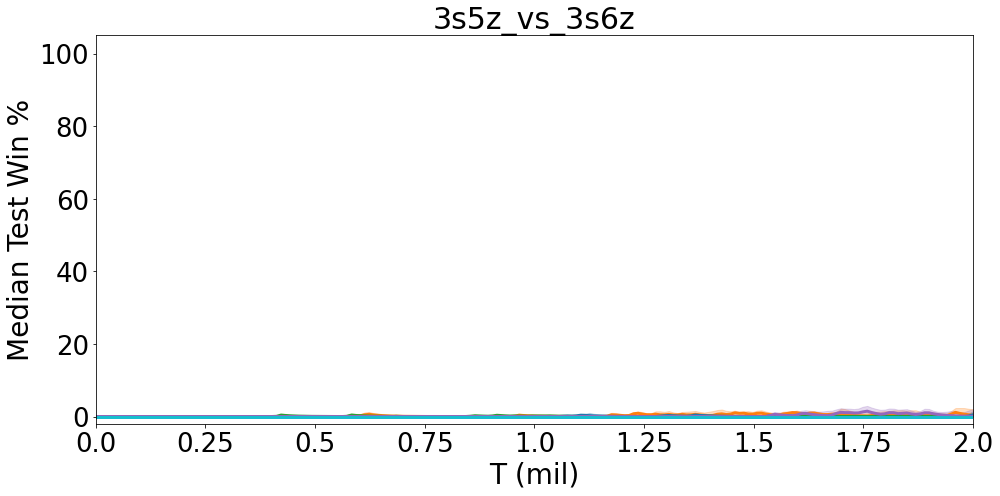}
   \end{subfigure}

   \begin{subfigure}[b]{0.3\textwidth}
      \centering
      \includegraphics[width=\textwidth]{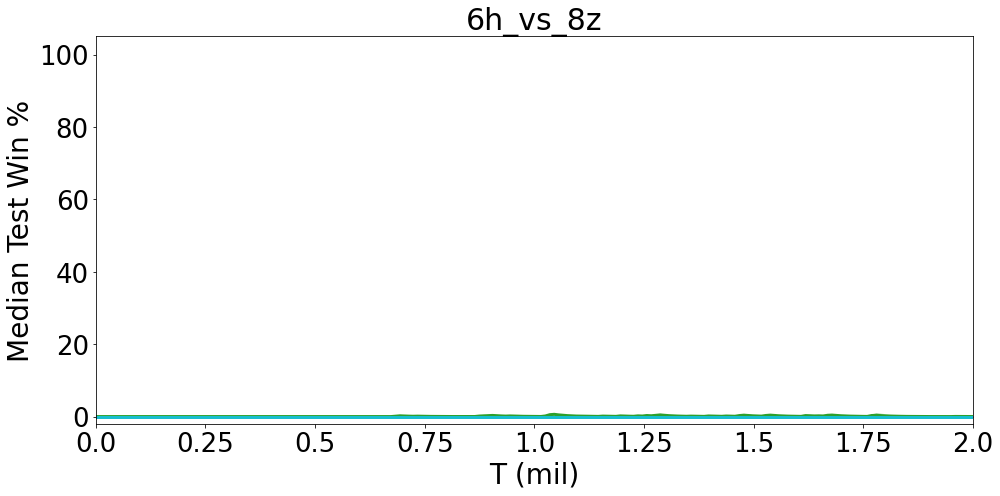}
   \end{subfigure}
   \hfill
   \begin{subfigure}[b]{0.3\textwidth}
      \centering
      \includegraphics[width=\textwidth]{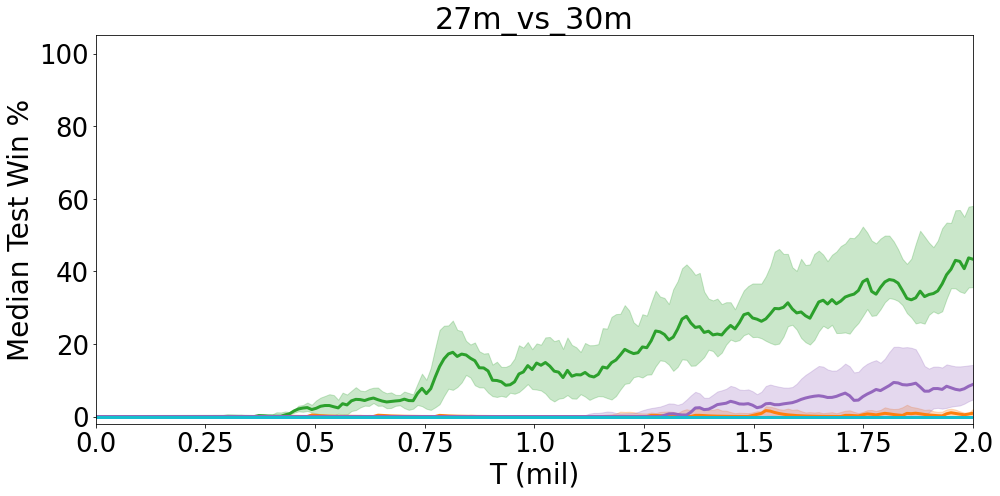}
   \end{subfigure}
   \hfill
   \begin{subfigure}[b]{0.3\textwidth}
      \centering
      \includegraphics[width=\textwidth]{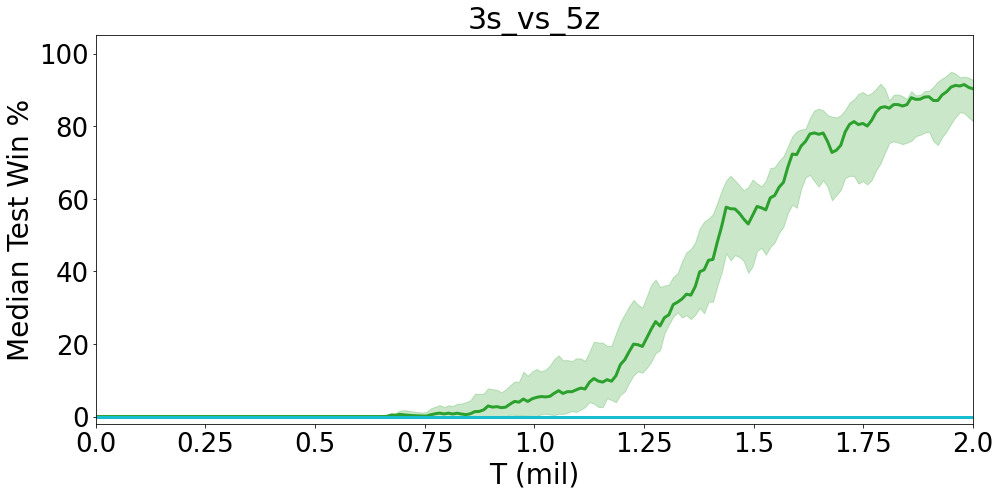}
   \end{subfigure}

   \begin{subfigure}[b]{0.3\textwidth}
      \centering
      \includegraphics[width=\textwidth]{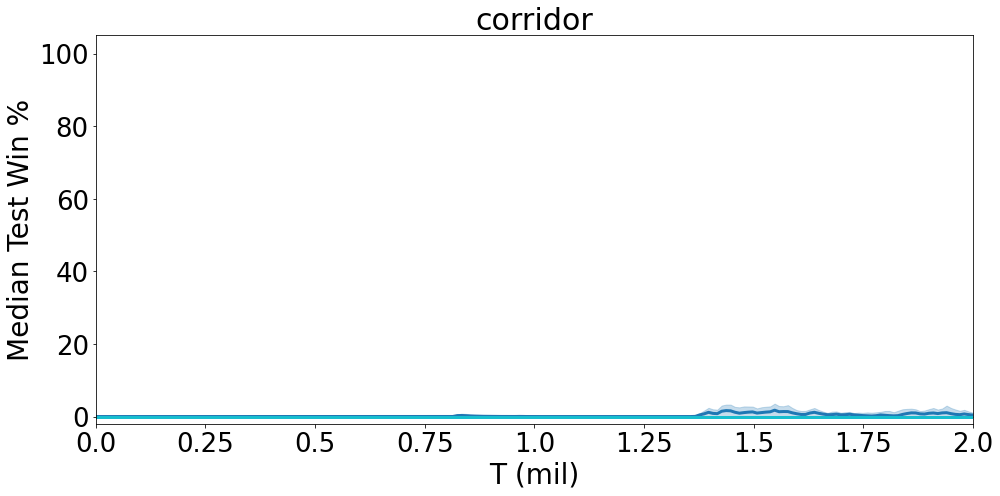}
   \end{subfigure}
   \hfill
   \begin{subfigure}[b]{0.3\textwidth}
      \centering
      \includegraphics[width=\textwidth]{ac_vs_qmix/MMM2_legend}
   \end{subfigure}
   \caption{Comparing MAPG methods with QMIX on the entire SMAC challenge. Batch size $32$.}
   \label{fig:ac_vs_qmix_full_results}
   \Description{Series of plots comparing CentralV-SOP, COMA-CC-SOP, QMIX, and DOP.}
\end{figure*}

The per-map comparisons of all algorithms we examine are presented in Figure \ref{fig:ac_vs_qmix_full_results}.

\subsection{CentralV-default}
\label{appx:centralv_default}
\begin{figure*}[t!]
   \centering
   \begin{subfigure}[b]{0.3\textwidth}
       \centering
       \includegraphics[width=\textwidth]{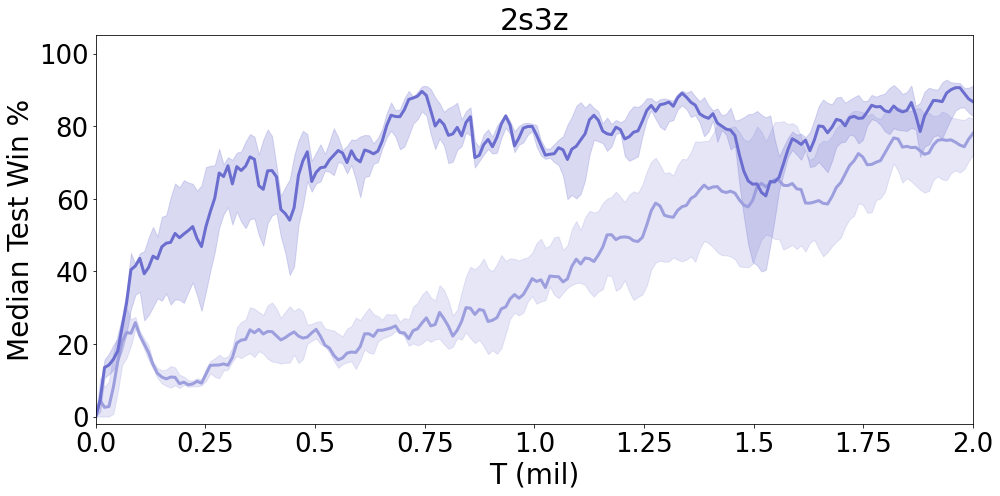}
   \end{subfigure}
   \hfill
   \begin{subfigure}[b]{0.3\textwidth}
       \centering
       \includegraphics[width=\textwidth]{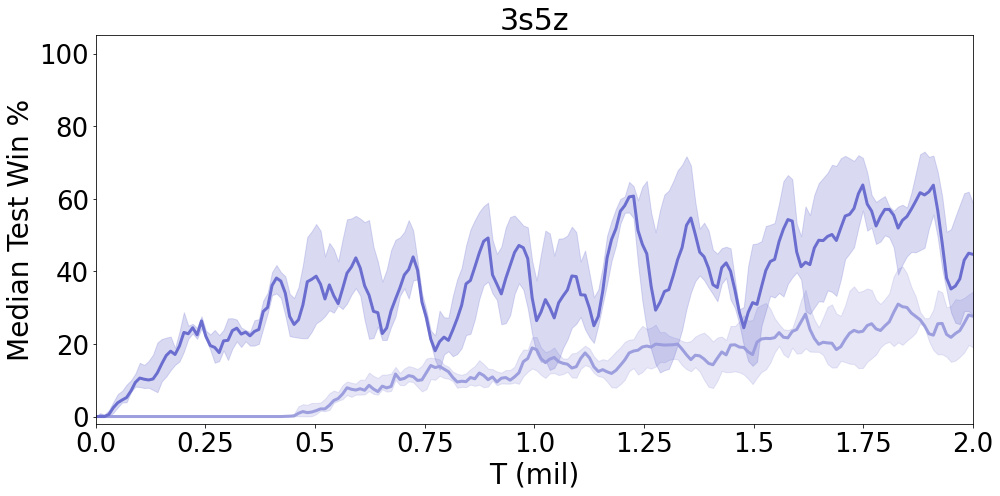}
   \end{subfigure}
   \hfill
   \begin{subfigure}[b]{0.3\textwidth}
       \centering
       \includegraphics[width=\textwidth]{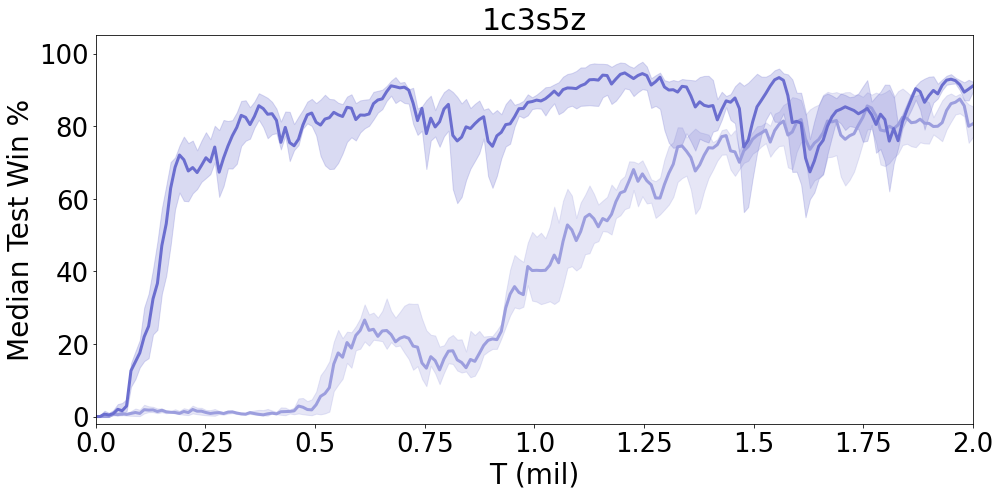}
   \end{subfigure}

   \begin{subfigure}[b]{0.3\textwidth}
      \centering
      \includegraphics[width=\textwidth]{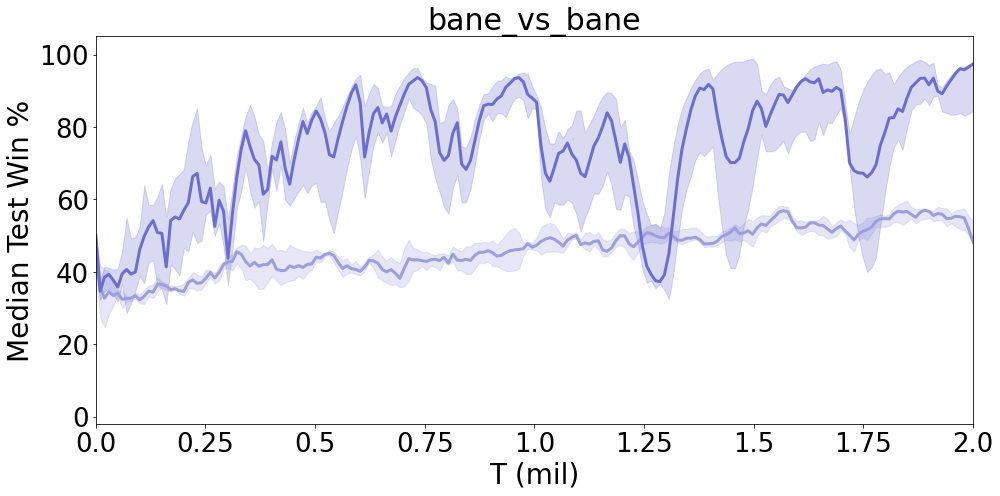}
  \end{subfigure}
  \hfill
  \begin{subfigure}[b]{0.3\textwidth}
      \centering
      \includegraphics[width=\textwidth]{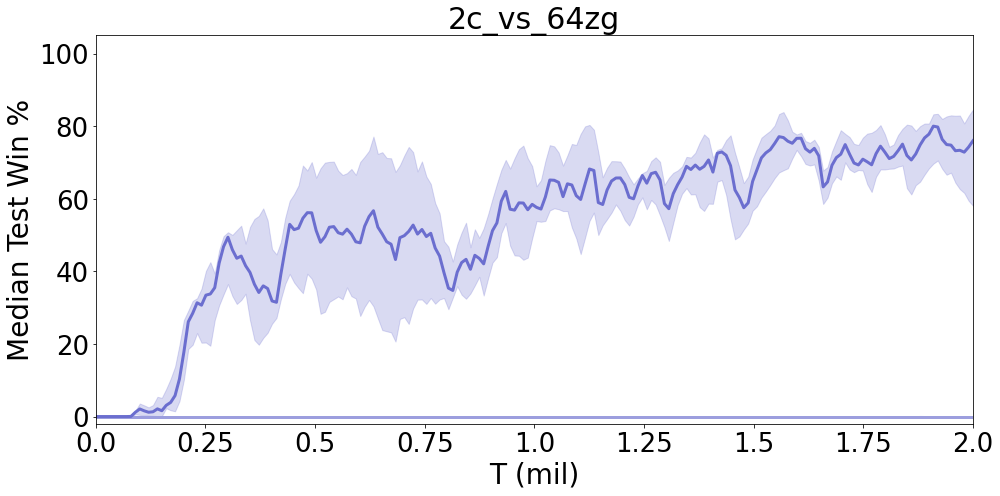}
  \end{subfigure}
  \hfill
  \begin{subfigure}[b]{0.3\textwidth}
      \centering
      \includegraphics[width=\textwidth]{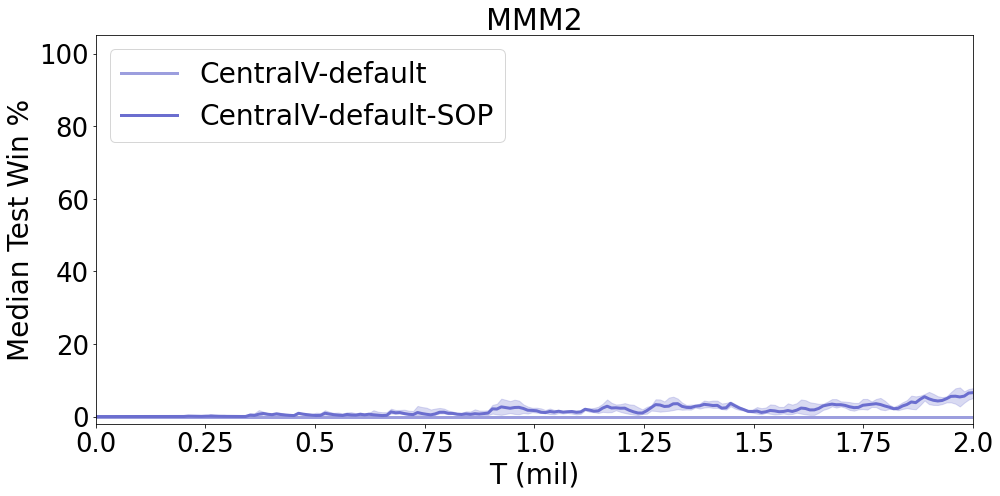}
  \end{subfigure}
 
   \caption{Comparing CentralV-default with SOP training and with purely on-policy training. Batch size $8$.}
   \label{fig:centralv_default_sop_vs_op}
   \Description{6 plots with titles 2s3z, 3s5z, 1c3s5z, bane\_vs\_bane, 2c\_vs\_64zg, MMM2. X-axis labelled T (mil). Y-axis labelled Median Test Win \%. Legend contains CentralV-default On-Policy Training and CentralV-default SOP Training.}
\end{figure*}

\begin{figure*}[t!]
   \centering
   \begin{subfigure}[b]{0.3\textwidth}
       \centering
       \includegraphics[width=\textwidth]{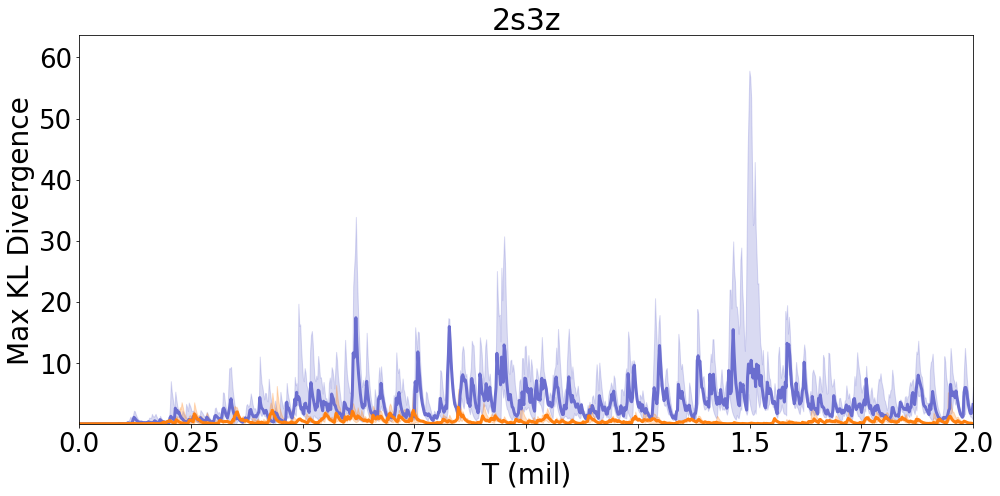}
   \end{subfigure}
   \hfill
   \begin{subfigure}[b]{0.3\textwidth}
       \centering
       \includegraphics[width=\textwidth]{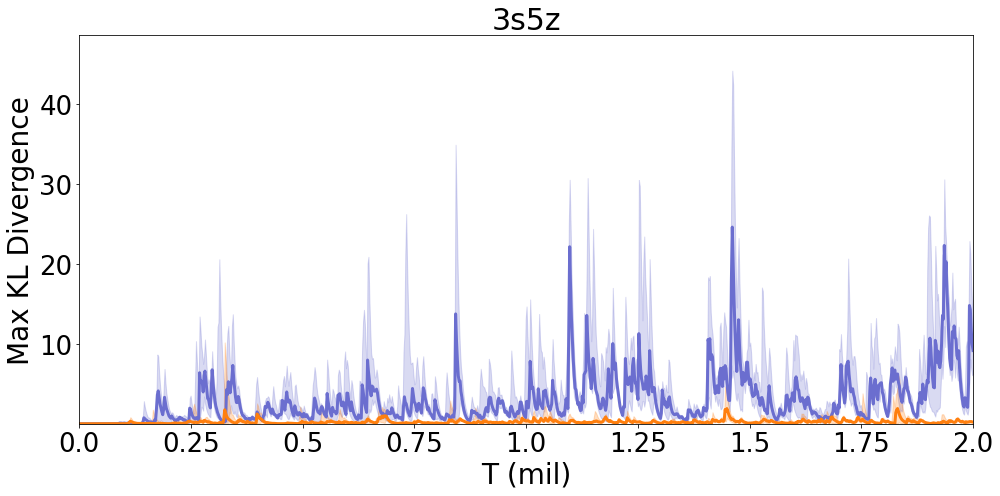}
   \end{subfigure}
   \hfill
   \begin{subfigure}[b]{0.3\textwidth}
       \centering
       \includegraphics[width=\textwidth]{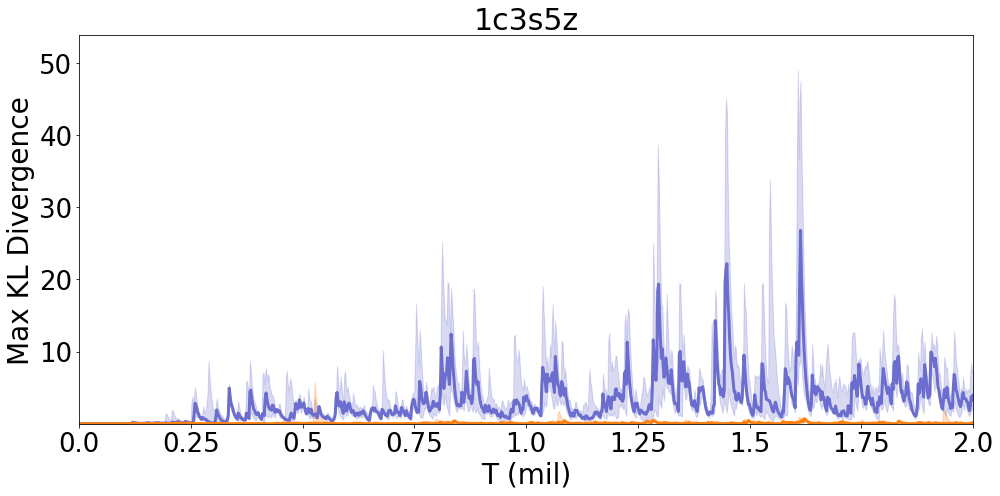}
   \end{subfigure}

   \begin{subfigure}[b]{0.3\textwidth}
      \centering
      \includegraphics[width=\textwidth]{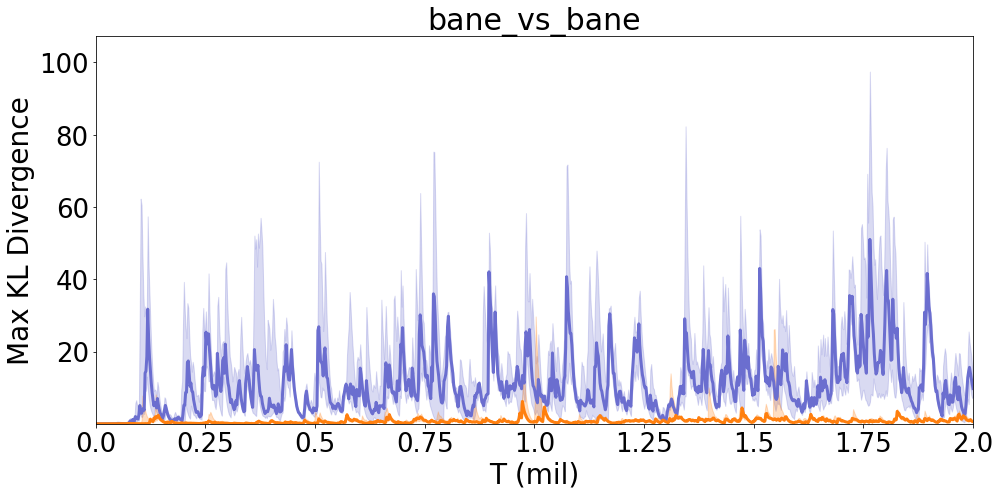}
  \end{subfigure}
  \hfill
  \begin{subfigure}[b]{0.3\textwidth}
      \centering
      \includegraphics[width=\textwidth]{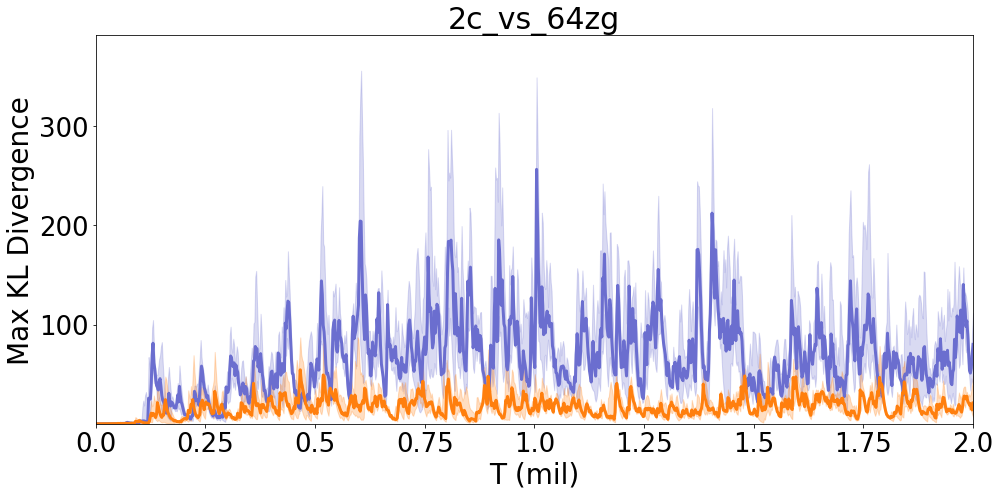}
  \end{subfigure}
  \hfill
  \begin{subfigure}[b]{0.3\textwidth}
      \centering
      \includegraphics[width=\textwidth]{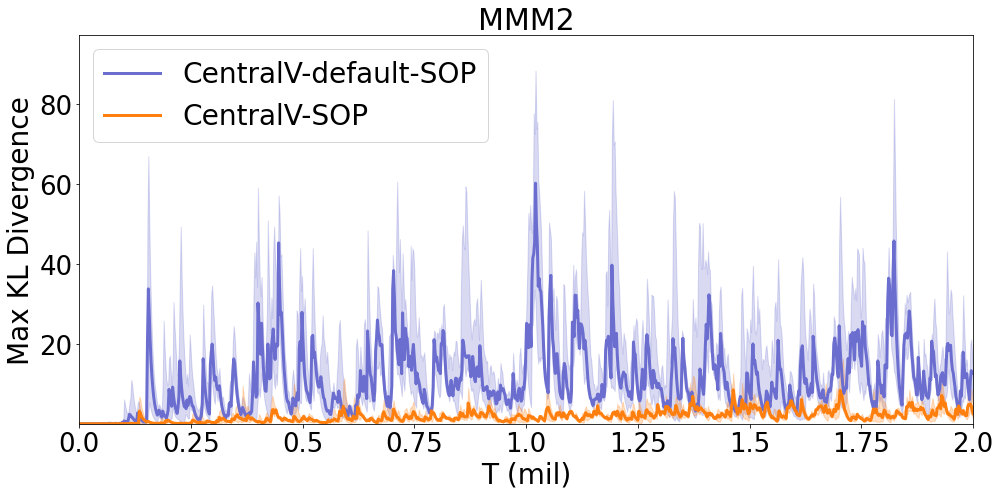}
  \end{subfigure}
 
   \caption{Comparing KL divergence of CentralV and CentralV-default with SOP training. Batch size $8$.}
   \Description{6 plots with titles 2s3z, 3s5z, 1c3s5z, bane\_vs\_bane, 2c\_vs\_64zg, MMM2. X-axis labelled T (mil). Y-axis labelled Max KL Divergence. Legend contains CentralV-SOP and CentralV.}
   \label{fig:centralv_sop_kl_comparison}
\end{figure*}

Here we show the results for using SOP training on CentralV without using the $\gamma = 1$ trick when computing advantages. We refer to this version as CentralV-default. The results are shown in Figure \ref{fig:centralv_default_sop_vs_op}. SOP training still manages to improve the performance of the algorithm. The increased training instability can be explained by unstable policy updates, as shown in Figure \ref{fig:centralv_sop_kl_comparison}.

\clearpage

\end{document}